\documentclass[10pt,journal,compsoc]{IEEEtran}

\usepackage[nocompress]{cite}
\usepackage{times}
\usepackage{epsfig}
\usepackage{graphicx}
\usepackage{amsmath}
\usepackage{amssymb}
\usepackage{dsfont}
\usepackage{calrsfs}
\usepackage{threeparttable}
\usepackage[linesnumbered,ruled,vlined]{algorithm2e}
\usepackage{algorithmicx}
\usepackage{xcolor}
\usepackage{mathtools}
\usepackage{array}
\usepackage{booktabs}
\usepackage{bm} 
\usepackage{amsfonts} 
\usepackage{multirow}
\usepackage{enumerate}
\usepackage{enumitem}
\usepackage{xspace}
\usepackage{makecell}
\usepackage[subnum]{cases}
\usepackage[caption=false,font=footnotesize]{subfig}
\usepackage{dsfont}
\graphicspath{ {IMAGES/} }
\usepackage{url}
\usepackage{balance}

\newcommand\tab[1][0.5cm]{\hspace*{#1}}

\def\ie{\textit{i.e.},\@\xspace}

\DeclarePairedDelimiter\floor{\lfloor}{\rfloor}           
\begin{document}

\title{Face Phylogeny Tree Using Basis Functions}
\author{Sudipta~Banerjee,~\IEEEmembership{Member,~IEEE}
        and~Arun~Ross,~\IEEEmembership{Senior~Member,~IEEE}%
        
\IEEEcompsocitemizethanks{\IEEEcompsocthanksitem Both the authors are with the Department
of Computer Science and Engineering, Michigan State University, East Lansing,
MI, 48824.\protect\\
E-mail: \{banerj24, rossarun\} @ cse.msu.edu}
}


\IEEEtitleabstractindextext{%
\begin{abstract}
Photometric transformations, such as brightness and contrast adjustment, can be applied to a face image repeatedly creating a set of near-duplicate images. Identifying the original image from a set of such near-duplicates and deducing the relationship between them are important in the context of digital image forensics. This is commonly done by generating an image phylogeny tree \textemdash \hspace{0.08cm} a hierarchical structure depicting the relationship between a set of near-duplicate images. In this work, we utilize three different families of basis functions to model pairwise relationships between near-duplicate images. The basis functions used in this work are orthogonal polynomials, wavelet basis functions and radial basis functions. We perform extensive experiments to assess the performance of the proposed method across three different modalities, namely, face, fingerprint and iris images; across different image phylogeny tree configurations; and across different types of photometric transformations. We also utilize the same basis functions to model geometric transformations and deep-learning based transformations.
We also perform extensive analysis of each basis function with respect to its ability to model arbitrary  transformations and to distinguish between the original and the transformed images. Finally, we utilize the concept of approximate von Neumann graph entropy to explain the success and failure cases of the proposed IPT generation algorithm. Experiments indicate that the proposed algorithm generalizes well across different scenarios thereby suggesting the merits of using basis functions to model the relationship between photometrically and geometrically modified images.



\end{abstract}

}

\maketitle

\IEEEdisplaynontitleabstractindextext

%
\IEEEpeerreviewmaketitle

\begin{figure}[h]
\centering
\subfloat[]
{
    \includegraphics[scale=.25]{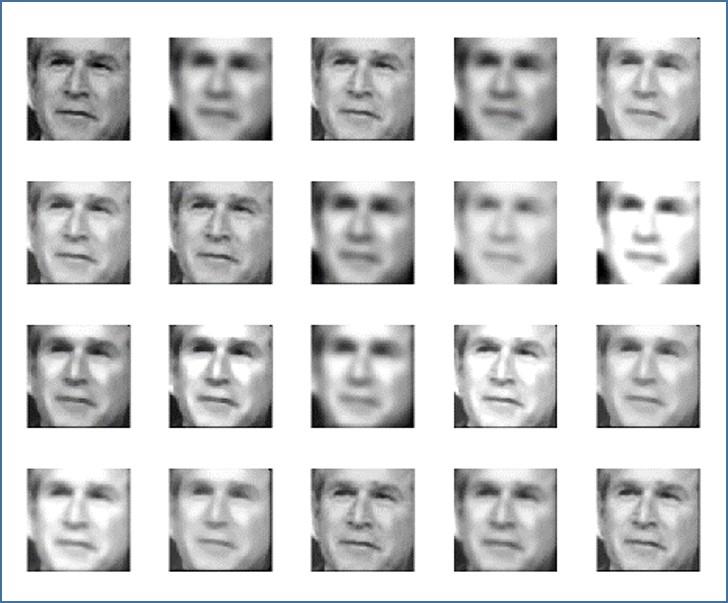} 
    }
 \subfloat[]
{
    \includegraphics[scale=.25]{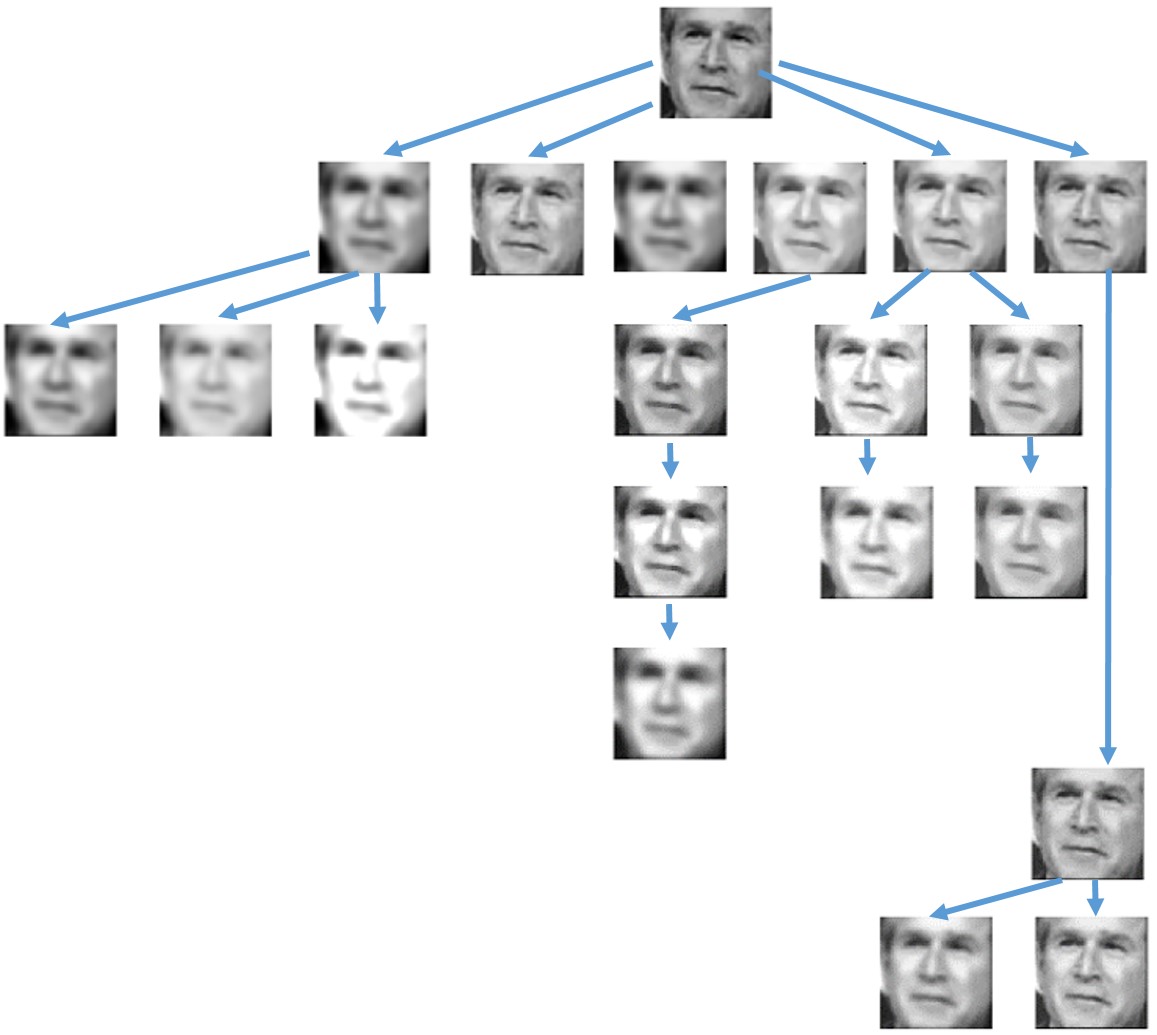} 
    }   
\caption{Examples of photometrically related near-duplicate face images. (a) A set of 20 related images and (b) their corresponding Image Phylogeny Tree (IPT). In our work (a) is the input and (b) is the output.}
\label{Fig:IPT_20}
\end{figure}

\ifCLASSOPTIONcompsoc
\IEEEraisesectionheading{\section{Introduction}\label{sec:introduction}}
\else
\section{Introduction}
\label{sec:introduction}
\fi
\IEEEPARstart{I}{n} many applications, the face image of an individual may be subjected to photometric transformations such as brightness adjustment, histogram equalization, gamma correction, etc. These photometric transformations may be applied in a sequential fashion, resulting in an array of near-duplicate face images (see Figure~\ref{Fig:JL}). While some of these transformations can be used to improve face recognition~\cite{Face}, others may be maliciously used for image `tampering'~\cite{Face_tampering}. The availability of inexpensive photo editing tools and the ability to rapidly share images across social media platforms, has resulted in the posting of a large number of near-duplicates on the internet. Identification of the original image from a set of such near-duplicates is important in the context of digital image forensics~\cite{Farid_08_DIF}. Further, since these images are created by the successive application of photometric transformations, inferring the \textit{order of evolution} is a challenging but interesting problem~\cite{Ross_ICB_19}. The order of evolution can be represented as an Image Phylogeny Tree (IPT), that indicates the relationship between the root node (original image) and the child nodes (transformed images) via directed links as illustrated in Figure~\ref{Fig:IPT_20}. Deducing the IPT from the set of near-duplicates in an automated fashion has the following advantages:

\begin{figure}[h]
\centering
\subfloat[]
{
    \includegraphics[scale=.14]{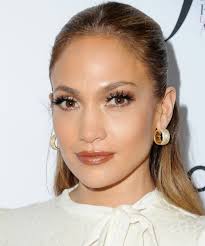} 
    } \hfill
 \subfloat[]
{
    \includegraphics[scale=.14]{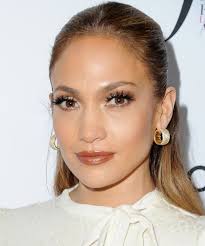} 
    } \hfill 
     \subfloat[]
{
    \includegraphics[scale=.14]{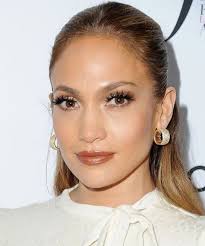} 
    } \hfill 
     \subfloat[]
{
    \includegraphics[scale=.14]{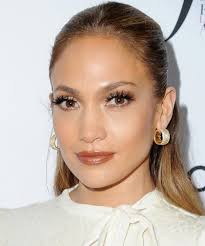} 
    } 
\caption{Examples of variations of the same image uploaded on multiple websites with subtle modifications making them appear almost identical.}
\label{Fig:JL}
\end{figure} 

\noindent \textbf{1. Indication of image tampering:} An image can be tampered for a number of reasons. It can be done to airbrush celebrity faces on magazine covers,\footnote{\url{https://www.cbsnews.com/news/uk-curb-airbrushed-images-keep-bodies-real/}} or to depict fake situations to garner political attention. In either case, tampered images convey false information. An IPT has directed links, and therefore, can be used to trace an image back to its origin, \ie the root node that denotes the original image. \\
\noindent \textbf{2. Preserving chain of custody:} Face images can be produced as culpable biometric evidence in legal proceedings~\cite{ImageForensics_FBI}. The admissibility of such evidence is contingent on its integrity, \ie it should not have been tampered with. The chain of custody~\cite{Chainofcustody_Ref2} of the digital evidence can be established via (i) hardware identification (the device used to acquire the biometric sample)~\cite{ChainofCustody_FP}, and (ii) analysis at the image level. IPT construction involves image-level analysis and can be leveraged to determine the authenticity between a pair of biometric samples \ie the original versus the tampered.

In this work, we tackle this challenging problem in the context of different biometric modalities (face, fingerprint and iris images) subjected to different types of photometric and geometric transformations resulting in IPTs of different configurations.

The remainder of the paper is organized as follows. Section~\ref{Sec:Rel_Work} discusses the related work in the context of image phylogeny literature. Section~\ref{Sec:Prop_Method} describes the proposed method used to construct the IPT. Section~\ref{Sec:Expts} describes the dataset and the experiments conducted to evaluate the proposed method. Section~\ref{Sec:Results} reports the results and the performance of the proposed algorithm. Section~\ref{Sec:Explan} describes the graph entropy based measure to assess the reconstructed IPTs. Finally, Section~\ref{Sec:Concl} concludes the paper.   

\section{Related Work}
\label{Sec:Rel_Work}
The task of near-duplicate detection and retrieval (NDDR) is closely related to this work. NDDR is a well-researched topic which involves finding semantically related images for a given query image. In some cases, a query image may be a composite of multiple donor images, \ie portions of different images are \textit{donated} to produce the composite. Provenance analysis methods~\cite{Moreira_18_Provenance} have been used to identify the donor images of a composite. In such cases, undirected phylogeny trees~\cite{Bharati_17_UPhylogeny} are suitable, since deducing the links is more critical than determining the direction of the link. Image phylogeny trees (IPTs), on the other hand, deal with the task of determining ancestral relationships, \ie explicitly determining the parent and child node as well as the predecessors. In the case of IPT construction, the typical assumption is that there is a single root node and all the images are related such that there is no isolated node~\cite{Dias_12_MST, Dias_13_largescaleIPT, Dias_13_OB, Bestagini_16_regionbasedIPT, Philippe_16_missingmarkersIPT, Milani_16_agemetricsIPT}. We make the same assumption in this paper. Other works exist that consider the presence of multiple root nodes resulting in Image Phylogeny Forests~\cite{Dias_13_AutoIPF, Oliveira_14_multipleparentingIPT, SpecClus_16_Dias}.   
 
The construction of an IPT commonly comprises of two major steps:
\begin{itemize}
\item Firstly, a pairwise asymmetric (dis)similarity measure is used to determine the degree of (dis)similarity between a pair of images and to discriminate successfully between the forward and reverse transformation directions. For an image pair $(\bm{I}_i, \bm{I}_j)$, where, $\bm{I}_i$ is the original image and $\bm{I}_j$ is the transformed image, the forward transformation direction refers to $\bm{I}_i \rightarrow \bm{I}_j$ and the reverse direction refers to $\bm{I}_j \rightarrow \bm{I}_i$. For a set of $n$ near-duplicate images, this pairwise asymmetric measure is used to populate an $n \times n$ matrix.
\item Secondly, a tree spanning algorithm utilizes the asymmetric matrix computed in the first step to determine the relationship between nodes and deduce a hierarchical tree structure.
\end{itemize}

In~\cite{Dias_12_MST}, the IPT is considered to be a minimal spanning tree (MST). In this work, we consider the ancestral links also as correct edges. Therefore, we interpret IPT as a directed acyclic graph (DAG) while relaxing the MST criterion. Majority of the literature focuses on different kinds of geometric transformations, \textit{e.g.}, cropping, scaling, rotation; and pixel intensity-based transformations, \textit{e.g.}, brightness and contrast adjustment, gamma transformation, and compression operations. Thus, traditionally, the first step of asymmetric measure computation involves geometric registration, followed by color channel normalization and compression matching~\cite{Dias_12_MST}. Other works focus on using wavelet-based denoising technique~\cite{Melloni_14_dissmetricsIPT}, and a combination of gradient estimation and mutual information techniques~\cite{Costa_17} to derive an improved asymmetric measure. 

The emphasis in the aforementioned works are on images depicting natural scenes. In this work, we focus on biometric images, \textit{viz.}, face, fingerprints and iris. Recently, work has been done in modeling photometric transformations for iris images~\cite{Ban_17_IJCB} using simple linear and quadratic functions. In~\cite{FacePhyloTree_2019}, Legendre polynomials and Gaussian Radial Basis Functions were employed to model photometric transformations for face images. The basis functions play an important role in not only reliably modeling the transformations, but also, in successfully discriminating between the forward and reverse directions. Therefore, the selection of a suitable set of basis functions is necessary to model the transformations and to distinguish between forward and reverse transformations. In the literature, there are a number of basis functions from the polynomial, wavelet and radial basis families. Even within a family, the basis functions can differ from each other, in terms of their ranges of orthogonality, associated weights, and other properties. In this paper, we expand on the work done in~\cite{FacePhyloTree_2019}, by considering a larger number of potential basis functions. Our work \textit{differs} from the previous work~\cite{FacePhyloTree_2019} as follows:

\begin{enumerate}
\item We consider three different families of basis functions for modeling photometric and geometric transformations: (i) Orthogonal polynomial family (Legendre and Chebyshev), (ii) Wavelet family (Gabor), and (iii) Radial Basis family (Gaussian and Bump).
\item We perform cross-modality testing, \ie learning the parameters of the basis functions using \textit{face} images, and testing it on near-infrared \textit{iris} images and optical sensor \textit{fingerprint} images.  
\item We test on multiple IPT configurations to evaluate the robustness of the proposed method. Also, we assess our method's robustness to unseen photometric and geometric transformations (\textit{i.e.,} transformations not used during the training phase) accomplished using deep learning-based schemes, as well as open-source and commercial software. Furthermore, we have performed qualitative assessment of the IPTs reconstructed using the proposed method on near-duplicates downloaded from the internet. 
\item We visualize the results using \textit{t}-distributed stochastic neighbor embedding (\textit{t}-SNE) to better understand the ability of the basis functions in modeling the transformations and discriminating between forward and reverse transformation directions. 
\item We employ von Neumann directed graph entropy to better understand and evaluate the reconstructed IPTs.
\end{enumerate}

\section{Proposed Method}
\label{Sec:Prop_Method}
A photometrically related image pair ($\bm{I}_i, \bm{I}_j$) can be generated by applying a single transformation or a sequence of transformations to one image resulting in the other image. However, to construct the IPT we require to differentiate between the original image and the transformed image. Say, if $\bm{I}_i$ is the original image and $\bm{I}_j$ is the transformed image, then the IPT should have a directed link as follows: $\bm{I}_i \rightarrow \bm{I}_j$. Applying this same principle to a set of near-duplicate photometrically related images, we need two sets: the first set denoting the \textit{parent} nodes and the second set denoting the \textit{child} nodes. These two sets are then used to construct the IPT ($ parent \rightarrow child $). So the first step is to identify the sets of parent and child nodes from an array of near-duplicates.

We proceed to identify the parent and the child node for each pair of images by first modeling the transformation that relates the two images. We use parameterized basis functions to model the transformations in both directions ($\bm{I}_i \rightarrow \bm{I}_j$ and $\bm{I}_i \leftarrow \bm{I}_j$). But modeling the transformation does not indicate which is the parent node and which is the child node. To accomplish this, we require an asymmetric measure to distinguish between the forward and reverse directions. We pose the asymmetric measure computation as the \textit{likelihood} ratio problem~\cite{FacePhyloTree_2019}. To compute the likelihood ratio, we adopt a supervised framework with a training phase and a testing phase. In the \textit{training} phase, we model numerous transformations for a large number of near-duplicate image pairs in both directions (in this phase we know the original and the transformed images \textit{apriori}). This results in two sets of parameter distributions, one for the forward transformation and the other for the reverse transformation. In the \textit{testing} phase, for a given near-duplicate pair, we first model the transformations in both directions. Next, we use the estimated parameters to determine how \textit{likely} they are to originate from the forward parameter distribution as opposed to the reverse parameter distribution. This step leads to the computation of the asymmetric similarity measure. We repeat this step for all image pairs in the near-duplicate set. Upon pairwise modeling of all the near-duplicate images in the set, we perform thresholding to identify \textit{related} image pairs. The similarity measure is then utilized to identify which image from the related pair is the parent, thus, making the other image its child. The sets of parent and child nodes are ultimately used to generate the IPT.

In this work, we seek to model an arbitrary transformation using a set of parametric functions, that we refer to as basis functions. Such an approach is needed since the space of photometric transformations is very vast; further, each of these transformations has a large number of parameter values. For example, a simple brightness adjustment can be accomplished using a large number of brightness values. The use of a fixed set of parametric functions to approximate a transformation reduces the otherwise complex task of modeling the photometric transformation. Thus, the task of approximating the transformations involves learning the \textit{parameters} of the basis functions, subject to a criterion. In our case, the criterion or the objective function is the minimization of the photometric error between a near-duplicate image pair. This is formulated as below:
\begin{equation}
\label{Eq:PE}
PE(\bm{I_i},\bm{I_j})=\min_{\bm{\alpha}}  \Sigma_p \|\bm{I}_i(p) - \mathcal{T}[\bm{I}_j(p)|\bm{\alpha}]\|^2_2.
\end{equation} 

Here, $\mathcal{T[\cdot|\bm{\alpha}]}$ denotes the photometric transformation. We model the transformation using the basis function as $\bm{I}_i(p) \approx \mathcal{T}[\bm{I}_j(p)|\bm{\alpha}]\approx \sum_{h=1}^m \alpha_h \mathbb{B}_h[\bm{I}_j(p)]$, where the transformation is applied to each pixel $p$. $\bm{\alpha} = [\alpha_1,\cdots,\alpha_m]^T$ is the parameter vector to be estimated and $m$ is the number of basis functions. In this work, we have five different types of basis functions, so the value of $m$ depends upon the choice of the basis function. Next, we describe the process of modeling the transformations using the basis functions and the parameter estimation routines. 

\subsection{Parameter Estimation of Basis Functions}
\label{Param}

\subsubsection{Orthogonal Polynomial Basis Functions}
\begin{enumerate}
\item \textbf{Legendre polynomials} are a class of orthogonal polynomials defined in the interval [-1, 1]. The Legendre polynomial of degree $n$ computed at $x$ is denoted as $P_n(x)$ and is written as follows:
\begin{equation}
L_n(x) = 2^n\sum_{k=0}^n x^k \binom{n}{k} \binom{\frac{n+k-1}{2}}{n}.
\label{Eq:Leg}
\end{equation}  
Legendre polynomials have been successfully used for image template matching~\cite{Leg1}, and image reconstruction and compression~\cite{Leg2}. Note that Eqn.(\ref{Eq:Leg}) simplifies to a linear function for $n=1$ and a quadratic polynomial for $n=2$.

\item \textbf{Chebyshev polynomials} are a special case of Jacobi polynomials defined in the interval [-1, 1]. There are two kinds of Chebyshev polynomials, here we are interested in Chebyshev polynomials of first kind which have been extensively used for approximating complex functions such as graph convolution~\cite{GCN_NIPS_16}. The explicit representation is presented below:
\begin{equation}
C_n(x) =  x^n\sum_{k=0}^{\floor*{\frac{n}{2}}} \binom{n}{2k} (1-x^{-2})^k.
\label{Eq:Cheb}
\end{equation} 

\end{enumerate}
In the notation, $\sum_{h=1}^m \alpha_h \mathbb{B}_h[\bm{I}_j(p)]$, $\mathbb{B}_h[\cdot]$ equals $L_h(\cdot)$ if Legendre polynomial is used and $C_h(\cdot)$ if Chebyshev polynomial is employed, and $m=n+1$. Next, we solve the objective function in Eqn.(\ref{Eq:PE}) using the inverse compositional estimation (ICE) algorithm~\cite{Bartoli_08_Photomodel, Baker_04_IJCV}. The IC update rule expresses the updated transformation as a composition of the current transformation and the inverse of the incremental transformation. See~\cite{Bartoli_08_Photomodel} for a detailed derivation of the IC update rule.
The parameter $\bm{\alpha}$ is computed as $\displaystyle \bm{\alpha} \leftarrow  \frac{\bm{\alpha}_{old}}{1+\Delta\bm{\alpha}}$. The IC update rule is an iterative optimization algorithm and updates the new $\bm{\alpha}$ using the previous value, $\bm{\alpha}_{old}$, and the incremental $\Delta\bm{\alpha}$. The incremental parameter vector is computed as,
\begin{equation}
\label{Eq:Param_OP}
\Delta\bm{\alpha}={({\bm{J}_{\bm{S}}}\bm{J}_{\bm{S}}}^T + \lambda\bm{Id})^{-1}{\bm{J}_{\bm{S}}}\bm{E}.
\end{equation}
Here, the term $\bm{J}_{\bm{S}}$ is known as the Jacobian of the source image ($\bm{I}_i$), and the term ${({\bm{J}_{\bm{S}}}\bm{J}_{\bm{S}}}^T)$ is known as the approximate Hessian matrix. We applied $L_2$ regularization to the Hessian matrix. Here, $\lambda$ denotes the regularization parameter, $\bm{Id}$ denotes the identity matrix, and $\bm{E}$ denotes the error image computed between the source image ($\bm{I}_i$) and the modeled target image ($\mathcal{T}[\bm{I}_j|\bm{\alpha_{old}}]$). In this work, $\bm{\alpha}$ is a 6-dimensional vector for both Legendre and Chebyshev polynomials.

\subsubsection{Wavelet Basis Functions}
\textbf{Gabor wavelets} are used for extracting texture information from images~\cite{Daug_Gabor} and has been selected as one of the basis functions for modeling the transformations. We employed a bank of Gabor filters parameterized with the wavelength and orientation. A set of four discrete wavelengths $\{2,3,4,5 \}$ and four orientations $\{0^{\circ}, 45^{\circ}, 90^{\circ}, 135^{\circ} \}$ are selected. Each wavelength corresponds to a single filter scale that treats the image at a different resolution. Thus, we have a bank of sixteen Gabor filters. We filtered the image with the Gabor bank and we obtained 16 filtered responses. However, in our case we combined the orientation responses for each wavelength, thus reducing the total number of responses from 16 to 4. Finally, we use ICE to estimate the 4-dimensional parameter vector $\bm{\alpha}$ ($m=4$).

\subsubsection{Radial Basis Functions} 
The polynomial and wavelet basis functions model the transformations at pixel level. In pixel-level modeling, the photometric error between each pixel of the original and transformed image pair is minimized using a weighted linear combination of basis functions. However, local filtering operations such as median filtering are applied in a patch-wise manner. Therefore, we used the family of radial basis functions that possesses nice smoothing properties to model transformations at the patch level. In patch-level modeling, the photometric error between two patches, one patch belonging to the original image and the second patch belonging to the transformed image, is minimized using a weighted linear combination of basis functions. Therefore, patch-level modeling considers all the pixels within a patch for minimizing the photometric error. This resolves the spatial dependencies observed in patch-based photometric transformations.
 
\begin{enumerate} 
\item \textbf{Gaussian radial} kernel is the first type of smoothing functions considered in the work. Gaussian RBF computed at $x$ is denoted as $K(x)$ and is written as $\displaystyle K(x)= \exp \| x-\mu\|^2$. Consider an image pair $(\bm{I}_i,\bm{I}_j)$ each of which is tessellated into $N_P$ non-overlapping blocks of size 16$\times$16. For the block $q$, where $q=[1,\cdots,N_P]$, let $\bm{I}_{j}^{q}(p) = \mathcal{T}[\bm{I}_{i}^{q}(p)|\bm{\alpha}_q]\approx \sum_{p} \alpha_{p,q} K[\bm{I}_{i}^{q}(p)]$. Here, $p$ denotes the pixel intensity value within the $q^{th}$ block and $\mu$ indicates the average of the pixel intensity values within that block. For Gaussian RBF, $m=p$, \ie for a 16$\times$16 block, the local least squares estimation yields $\bm{\alpha}_q$. Simplifying using the matrix notation yields $\bm{I}_{j}^{q}\approx \bm{\alpha}_q^T \bm{\mathbb{B}[I}_{i}^{q}]$, where $\bm{\mathbb{B}[I}_{i}^{q}] = K[\bm{I}_{i}^{q}(p)], \forall p$. The local least squares method is used to estimate the coefficient vector $\bm{\alpha}_q$ for each block. The final $\bm{\alpha}$ is a 256-dimensional vector obtained by computing the average of all $\bm{\alpha}_q$s. See~\cite{FacePhyloTree_2019} for detailed derivation.

\item \textbf{Bump RBF} is a smooth compact function which can be interpreted as a Gaussian function scaled to lie on a disc of unity radius. It is not analytic unlike Gaussian RBFs, but can be used as generalized functions which are essential in converting discontinuous functions to smooth functions~\cite{Bump}. In this case, $\displaystyle K(x)= \exp \bigg( -\frac{1}{1-x^2} \bigg)$ for $x \in [-1,1]$. Here, $x$ is mean-centered. Using least squares estimation, we obtain $\bm{\alpha}$ (256-dimensional).
\end{enumerate}

\subsection{Asymmetric Measure Computation and IPT Construction} 
\label{Diss}

The asymmetric measure can be in the form of pairwise similarity or dissimilarity, but in this work, we adopt a similarity-based asymmetric measure. The similarity measure computed between a pair of images determines whether an image pair is photometrically related or not, \ie whether a link should exist between a pair of nodes (images) in the IPT; it also helps determine the direction of the link by identifying the parent node and the child node. The parameters estimated for modeling the transformations are utilized to compute this similarity measure as described below.

\subsubsection{Likelihood ratio for computing the asymmetric similarity measure}
\label{Assym}

Given a pair of images, $(\bm{I}_i, \bm{I}_j)$, we first estimate the parameter vectors $\bm{\alpha}_{ij}$ and $\bm{\alpha}_{ji}$ in both directions ($\bm{I}_i \rightarrow\bm{I}_j$ and  $\bm{I}_j \rightarrow\bm{I}_i$). The parameter vectors are necessary but not sufficient for constructing the IPT. We compute the \textit{likelihood ratio} from the estimated parameters to yield a similarity score which can discriminate between the forward and reverse directions, and can thus be used to construct the IPT. To compute the likelihood ratio, we need the probability distribution of the parameter vectors \textemdash \hspace{0.2cm} $p_f(\bm{\alpha})$ and $p_b(\bm{\alpha})$ corresponding to forward and reverse directions for a large number of training images. The probability distributions are generated in a supervised fashion, where we assume that we know for an image pair from the training set $(\bm{I}_r, \bm{I}_s)$, $\bm{I}_r$ is the original image and $\bm{I}_s$ is the transformed image. Then the forward transformation refers to $(\bm{I}_r \rightarrow\bm{I}_s)$, and the reverse transformation refers  to $(\bm{I}_s \rightarrow\bm{I}_r)$. The set of $\bm{\alpha}_{rs}$ vectors computed for a large number of image pairs are used to estimate $p_f(\bm{\alpha})$. Similarly, the set of $\bm{\alpha}_{sr}$ parameter vectors are used to determine $p_b(\bm{\alpha})$. We utilized Parzen window based non-parametric density estimation scheme~\cite{KDE} to obtain $p_f(\bm{\alpha})$ and $p_b(\bm{\alpha})$. 

Upon obtaining the forward and the reverse parameter distributions, we now compute the likelihood ratios as follows:
$\displaystyle \Lambda_{ij} = \frac{p_f(\bm{\alpha_{ij}})}{p_b(\bm{\alpha_{ij}})}$. Similarly, $\displaystyle \Lambda_{ji} = \frac{p_f(\bm{\alpha_{ji}})}{p_b(\bm{\alpha_{ji}})}$. Our intuition is that we will observe a higher value of $\Lambda_{ij}$ compared to $\Lambda_{ji}$, if $\bm{I}_i$ is the original image and $\bm{I}_j$ is the transformed image. In this case, $\bm{\alpha_{ij}}$ belongs to the forward distribution and should result in a higher value of $\Lambda_{ij}$. Conversely, $\bm{\alpha_{ji}}$ belongs to reverse distribution, resulting in a lower value of $\Lambda_{ij}$. The likelihood ratios are further used to populate the similarity matrix. 

\begin{figure}
\centering
    \includegraphics[scale=.23]{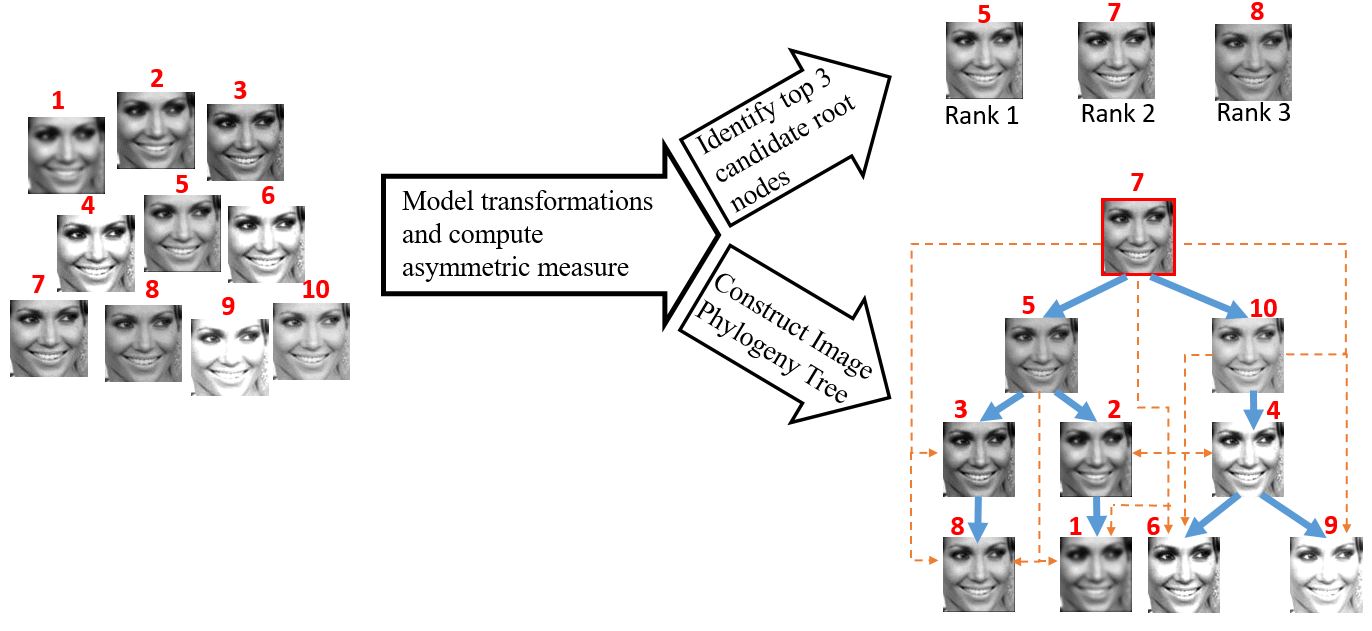} 
    
\caption{The outline of the proposed method. The proposed method first models the photometric transformations between every image pair and then computes the asymmetric measure. Given a set of near-duplicate images as input (on the left) the two objectives are: (i) to determine the candidate set of root nodes, and (ii) to construct the IPT when the root image is known. The dashed arrows indicate ancestral links and the bold arrows indicate immediate links between parent and child nodes. }
\label{Fig:Outline}
\end{figure} 

\subsubsection{IPT Construction}

\label{IPT}
The similarity matrix $\bm{S}$ of size $n \times n$ is populated by the likelihood ratio values as follows: $\bm{S}_{ij} = \Lambda_{ij}; i,j=[1,\cdots,n]$ and $i\neq j$. The diagonal elements of the similarity matrix are ignored as we do not consider self-loops in an IPT. The similarity matrix is then employed for (i) identifying the candidate root nodes and (ii) constructing the IPT. The steps are described below. 

\begin{enumerate}

\item \textbf{Indicator matrix representation:} This step helps in pruning the outliers which can be falsely identified as root nodes. The indicator matrix is constructed by thresholding the similarity matrix against a suitable threshold, which results in a binary matrix. The details of the threshold selection are described in~\cite{FacePhyloTree_2019}. The indicator matrix serves as an adjacency matrix of a coarse directed acyclic graph that is further refined for constructing the IPT. The reason it is referred to as ``coarse'' is that it may contain some spurious edges. 

\item \textbf{Candidate root nodes identification:} In this work, we consider each IPT to have a single root node. The authors in~\cite{Dias_13_AutoIPF, Dias_13_OB} considered each node as a potential root, one at a time, and then computed a cost function for each IPT constructed using the potential root. The IPT resulting in the least cost function value was selected, and its root node was used for the final evaluation. This involves $O(n^3)$ (can be optimized to $O(n^2)$) complexity as reported in~\cite{Dias_13_OB}. In contrast, the method proposed here computes a set of three candidate root nodes, which corresponds to the top 3 choices for the root node out of $n$ nodes. This requires finding the nodes having the highest number of 1's in the indicator matrix (we consider ancestral edges as correct edges). The entire process requires summing each row of the indicator matrix followed by sorting and this results in $O(n \log n)$ computational complexity.

\item \textbf{IPT generation:} We construct the IPT as described in~\cite{FacePhyloTree_2019} using a depth-first search-based tree spanning technique. The choice of depth-first search (DFS) over breadth-first search (BFS) is motivated by the fact that DFS has a linear memory requirement with respect to the nodes and results in a faster search ($O(n)$) and is therefore used for topological sorting. The total computational complexity for the IPT construction using the proposed method is $O(n \log n)+O(n) \approx O(n \log n)$.

\end{enumerate}

The outline of the proposed method for constructing the IPT is illustrated in Figure~\ref{Fig:Outline}.

\begin{table}[h!]
\centering
\caption{Description of the datasets used in this work.}
\label{Tab:Data}
\scalebox{0.88}{
\begin{tabular}{|lllll|} \hline
Modality                     & \begin{tabular}[c]{@{}l@{}}Name of the\\ Dataset\end{tabular}           & \begin{tabular}[c]{@{}l@{}}Dataset \\ Identifier\end{tabular} & \begin{tabular}[c]{@{}l@{}}No. of \\ subjects\end{tabular} & \begin{tabular}[c]{@{}l@{}}No. of \\ images\end{tabular} \\ \hline \hline
\multirow{2}{*}{Face}        & \multirow{2}{*}{LFW}                                                    & Partial Set                                                   & 391                                                        & 12,290                                                   \\
                             &                                                                         & Full set                                                      & 468                                                        & 27,270                                                   \\ \hline

\multirow{2}{*}{Iris} & \begin{tabular}[c]{@{}l@{}}CASIA-IrisV2\\ Device2\end{tabular}  &  \textemdash & 37 & 7,260                                                   \\
                             &   \begin{tabular}[c]{@{}l@{}}CASIA-IrisV4\\ Thousand\end{tabular}                                                                       &            \textemdash                                          & 525                                                         & 5,005
\\ \hline
\multirow{2}{*}{Fingerprint} & \multirow{2}{*}{\begin{tabular}[c]{@{}l@{}}FVC 2000\\ DB3\end{tabular}} & Config I                                                      & 110                                                        & 8,800                                                    \\
                             &                                                                         & Config II                                                     & 90                                                         & 7,200                                                   \\ \hline
\end{tabular}}
\end{table}

\begin{table*}[]
\centering
\caption{Photometric transformations and the range of the corresponding parameters used in the training and testing experiments. The transformed images are scaled to $[0,255]$. Note that experiments were also conducted using other complex photometric transformations besides the ones listed here.}
\label{Tab1:Params}

\begin{tabular}{|l|l|l|l|}
\hline
\textbf{Photometric Transformations} & \textbf{\begin{tabular}[c]{@{}l@{}}Level of \\ Operation\end{tabular}} & \textbf{Parameters}               & \textbf{Range}                                                      \\  \hline \hline
Brightness adjustment                & Global                                                                 & [a,b]                             & a $\in$ [0.9,1.5], b $\in$ [-30,30]                                            \\
Median filtering                     & Local                                                                  & size of window [m,n]              & m $\in$ [2,6], n $\in$ [2,6]                                              \\
Gaussian smoothing                   & Global                                                                 & standard deviation                & stddev $\in$[1,3]                                                      \\
Gamma transformation                     & Global                                                                 & gamma                             & gamma $\in$ [0.5,1.5]  \\ \hline                                                  
\end{tabular}
\end{table*}
  

\section{Experiments}
\label{Sec:Expts}
In this section, we describe the datasets employed, the experiments conducted and finally report the results.  
\subsection{Datasets}
We used four datasets belonging to three different modalities to conduct experiments. For the face modality, we used images from the Labeled Faces in the Wild (LFW) dataset~\cite{LFWTech}. For the iris modality, we used near-infrared iris images from the CASIA-IrisV2 Device2 subset~\cite{CasV2} and CASIA-IrisV4 Thousand subset~\cite{CASv4}. For the fingerprint modality, we used images from the FVC2000 DB3 dataset~\cite{FVC}. The description of the datasets is provided in Table~\ref{Tab:Data}. We selected four photometric transformations, \textit{viz}., Brightness adjustment, Median filtering filtering, Gaussian smoothing, and Gamma transformation as used in~\cite{FacePhyloTree_2019} to test the proposed IPT construction algorithm. The parameter range for each of the transformations is described in Table~\ref{Tab1:Params}.

\begin{figure*}[h]
\centering
\subfloat[]
{
    \includegraphics[scale=.31]{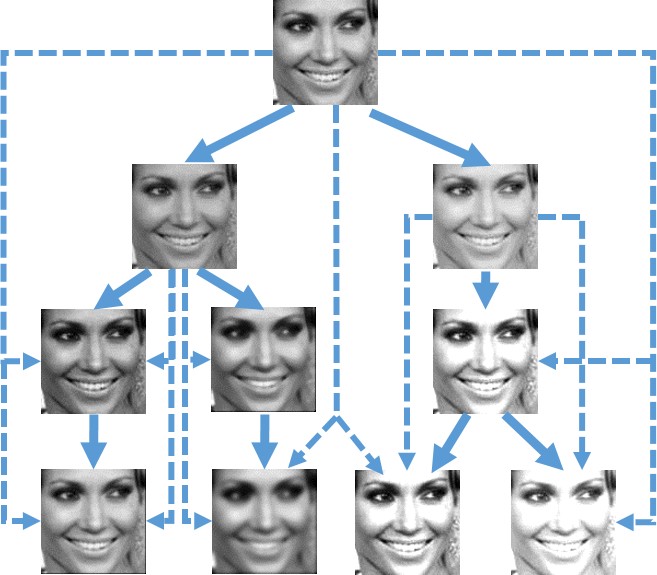} 
} \hfill
\subfloat[]
{
    \includegraphics[scale=.31]{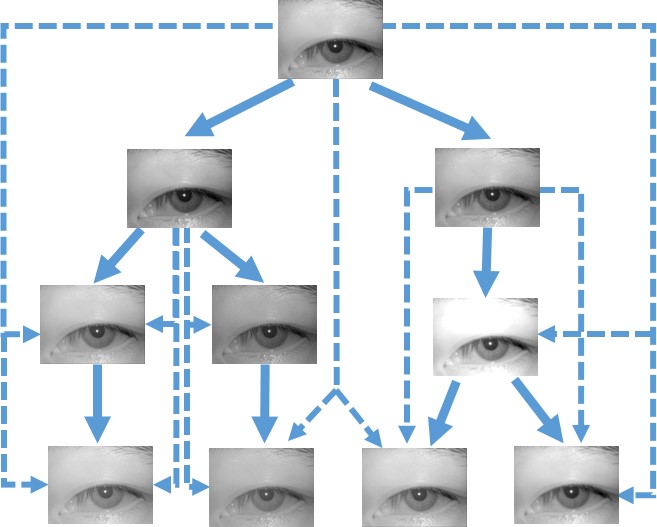} 
}\hfill
\subfloat[]
{
    \includegraphics[scale=.31]{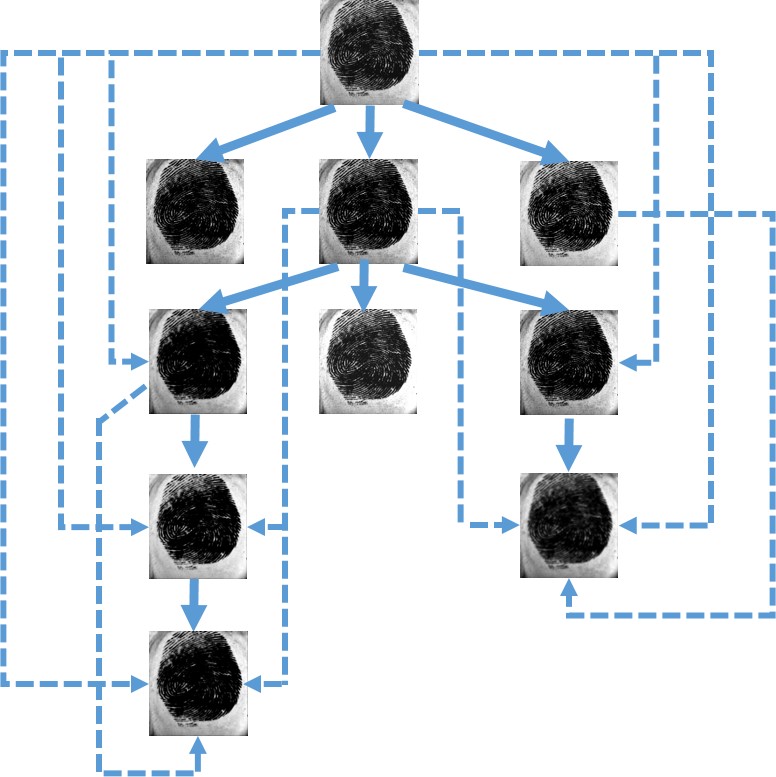} 
}\hfill
\subfloat[]
{
    \includegraphics[scale=.31]{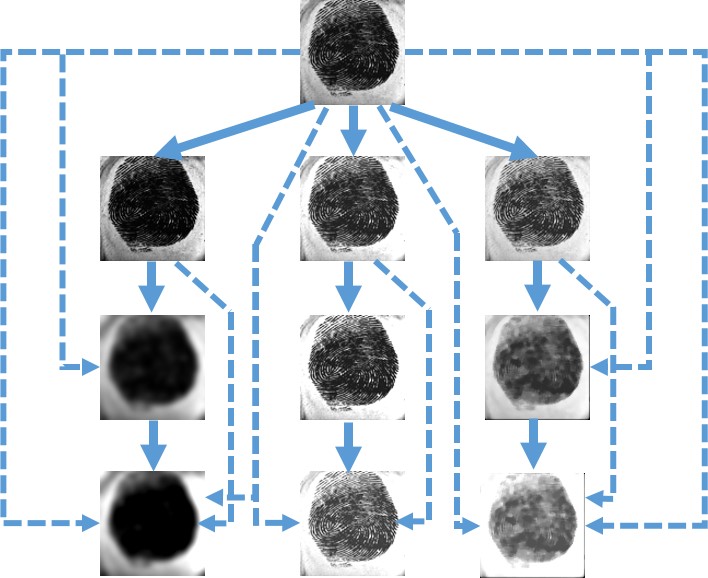} 
}

\caption{IPT configurations used in Experiments 2 and 3 for the face, iris and fingerprint modalities. Note that the same configuration was tested across two modalities (Face and Iris) while, two different configurations were tested for the same modality (Finger). The bold arrows indicate immediate links and the dashed arrows indicate ancestral links.}
\label{Fig:TestIPTs_ALL}

\end{figure*}

\begin{figure}
\centering
\subfloat[]
{
    \includegraphics[scale=.44]{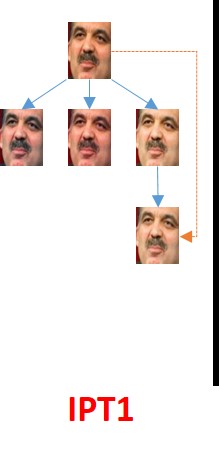}
    \label{fig:Val1}
}
\subfloat[]
{
    \includegraphics[scale=.44]{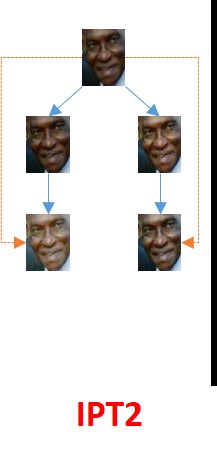}
    \label{fig:Val2}
}
\subfloat[]
{
    \includegraphics[scale=.44]{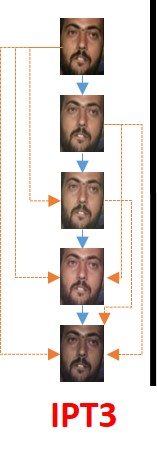}
    \label{fig:Val3}
}
\subfloat[]
{
    \includegraphics[scale=.44]{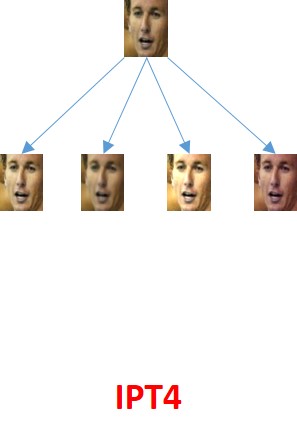}
    \label{fig:Val4}
}
\caption{IPT configurations used in Experiment 4. The bold arrows indicate immediate links and the dashed arrows indicate ancestral links.}
\label{Fig:Photoshop}
\end{figure}

\subsection{Experimental Methodology}
We performed seven experiments which are described below. 

\subsubsection{Experiment 1: Efficacy of basis functions}
In this experiment, we evaluate the ability of the basis functions to (i) model the photometric transformations and (ii) discriminate between the forward and reverse directions. To accomplish the first task, we perform two evaluation methods. The first evaluation involves \textit{deterministically} selected parameters, whereas the second evaluation involves \textit{randomly} selected transformation parameters. For the first evaluation, we select a face image, $\bm{I}$ and subject it to a single transformation, \textit{e.g.}, gamma adjustment, parameterized with a specific $\gamma$ value, resulting in $\bm{I'}$. We repeat this process 200 times, each time we use an incrementally modified $\gamma_{new}$ ($\gamma_{new} = \gamma_{old} + \Delta \gamma$), thus, resulting in 200 near-duplicate image pairs. Furthermore, we repeat this process for 5 images corresponding to 5 different subjects. Therefore, we have a total of 1,000 photometrically related image pairs for a single transformation. We conduct this process for each of the four transformations indicated in Table~\ref{Tab1:Params}. Next, we use the basis functions to model the transformation \textit{only} in the forward direction in this experiment. Then, we use $t$-SNE~\cite{TSNE} to reduce the dimensionality of the estimated vectors and project them onto 3-dimensions. This experiment is conducted to assess the ability of the basis functions in modeling the transformations and we visually interpret the results. The second protocol involves modeling 2,000 image pairs in the forward direction using the five basis functions, where each of the 500 images were subjected to each of the four photometric transformations using randomly selected parameters. We further computed the residual photometric error (PE), for each image pair which is the difference in the pixel intensity between the actual target image and the output modeled using the basis functions. We then computed the mean over all 2,000 image pairs. The mean of the residual PE evaluates the ability of the basis functions to accurately model the transformations.

To evaluate the second task, \ie discriminating between the forward and reverse directions, we selected 400 image pairs (100 image pairs corresponding to each of the four transformations) from the 2,000 pairs, that were generated using randomly selected parameters. Then, we modeled the transformations in \textit{both} forward and reverse directions, and estimated the parameters. We further used $t$-SNE to obtain a 2-dimensional embedding of the estimated parameters, and visualized the projections to analyze the performance of the basis functions.

\subsubsection{Experiment 2: IPT Reconstruction}
In this experiment, we evaluated the proposed approach in terms of (i) Root identification and (ii) IPT Reconstruction accuracy metrics as used in~\cite{FacePhyloTree_2019}. We followed the same experimental protocol in~\cite{FacePhyloTree_2019}, and assessed the performance of the basis functions on both a partial set and a full-set of face images from the LFW dataset.\footnote{\url{http://iprobe.cse.msu.edu/dataset_detail.php?id=1&?title=Near-Duplicate_Face_Images_(NDFI)}} We used the IPT configuration presented in Figure~\ref{Fig:TestIPTs_ALL}(a) for this experiment. 

\subsubsection{Experiment 3: Cross-modality testing on multiple configurations} 
\label{subsubsect:expt3}
We tested the proposed approach on iris images and fingerprint images. This experiment is intended to demonstrate the generalizability of the proposed method across modalities. 

\begin{enumerate}
\item \textbf{Iris Images} \textemdash We applied a random sequence of photometric transformations on near-infrared iris images. We evaluated the root identification and IPT reconstruction accuracies for 726 IPTs using the same procedure as described in Section~\ref{Diss}. We used the same IPT configuration as the one used for face images (see Figure~\ref{Fig:TestIPTs_ALL}(b)). Note, the parameter probability distributions in the forward and reverse directions are computed using a training set comprising of \textit{face} images; the test images are \textit{iris} images. The test iris images are acquired in the near-infrared spectrum, in contrast to the training face images that are acquired in the visible spectrum. As a result, this experiment can also be treated as an assessment of the basis functions for cross-spectral modeling. 

\item \textbf{Fingerprint Images} \textemdash Two different IPT configurations are tested as depicted in Figures~\ref{Fig:TestIPTs_ALL}(c) and \ref{Fig:TestIPTs_ALL}(d). This experiment tests the generalizability of the proposed approach as a function of the breadth and depth of the IPT. We refer to Figure~\ref{Fig:TestIPTs_ALL}(c) as Config I and Figure~\ref{Fig:TestIPTs_ALL}(d) as Config II. The IPT configuration used for testing the face and iris images is more balanced (similar distribution of nodes on the left and the right sides of the root) compared to the Config I structure which has more depth than breadth, whereas Config II has the same breadth at successive depths. 
\end{enumerate}

We also performed an intra-modality experiment which serves as the baseline experiment to compare against the performance of the cross-modality experiment. The train and test partitions for the intra-modality experiments are as follows:
\begin{itemize}

\item Iris images \textemdash \hspace{0.12cm} We used 5,005 images from the CASIA-IrisV4 Thousand subset belonging to 525 subjects to learn the parameter distributions in the forward and the reverse directions. We then tested it on 726 IPTs (same configuration as in Figure~\ref{Fig:TestIPTs_ALL}(b)) constructed from the CASIA-IrisV2 Device2 subset. This experiment can also be considered as a \textit{cross-dataset} experiment, due to the use of two different datasets in the training and testing phases.

\item Fingerprint images \textemdash \hspace{0.12cm} We used the same dataset (\textit{intra-dataset}) in training and testing but we strictly followed a subject disjoint protocol. This, however, resulted in a lesser number of training images. 560 images from 70 subjects were used for creating parameter distributions and then tested on 3,200 images from 40 subjects. We used the same configurations as depicted in Figures~\ref{Fig:TestIPTs_ALL}(c) and~\ref{Fig:TestIPTs_ALL}(d).

\end{itemize}
 
\subsubsection{Experiment 4: Robustness to unseen photometric transformations} We considered a closed set of 4 transformations in the training stage. However, a gamut of image and video editing tools such as Photoshop, GIMP and Snapchat filters exist which can be used for image manipulation, particularly for face images. In this context, we constructed a small test set of images transformed using Photoshop operations (Hue and Saturation adjustment, Curve transformation, Color balance, and Blur filters) to create 35 IPTs corresponding to 5 subjects. The IPT configurations are selected such that they cover diverse breadth and depth values possible for an IPT with 5 nodes. See Figure~\ref{Fig:Photoshop}. The trained parameter distributions did not encounter instances of Photoshopped images; hence, this experiment will demonstrate the robustness of the basis functions in handling unseen transformations.

\subsubsection{Experiment 5: Ability to handle geometric transformations}
\begin{figure}
\centering

   \includegraphics[scale=.28]{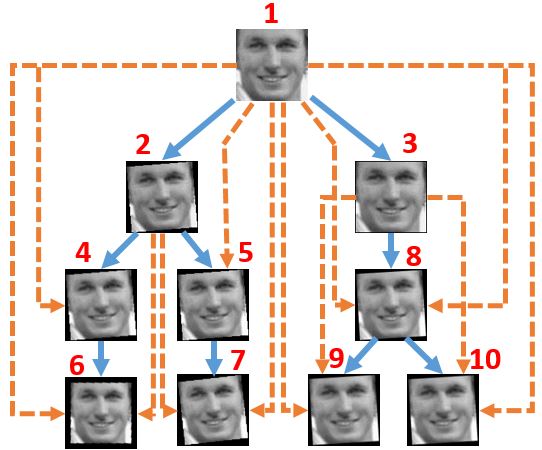}\hfill
 
\caption{Experiment 5: An example IPT generated using geometrically modified near-duplicate images. The bold arrows indicate immediate links and the dashed arrows indicate ancestral links.}
\label{Fig:Geom_IPT}
\end{figure}

We designed this experiment to assess the ability of basis functions in modeling geometric transformations. We selected some well-known geometric transformations such as sampling using linear interpolation and affine transformations that include translation, scaling and rotation. We have selected these particular transformations as they have also been utilized in~\cite{Dias_12_MST} for creating near-duplicates. The details about the geometric transformations and their respective parameter ranges are described in Table~\ref{Tab:GeomParams}. We randomly selected 500 images belonging to 97 subjects from the Labeled Faces in the wild (LFW) dataset~\cite{LFWTech}. We then applied four geometric modifications (see Table~\ref{Tab:GeomParams}) on each of these images in a random sequence with random parameter values to create 500 image phylogeny trees (IPTs). Each IPT contains 10 images so we have a total of 5,000 images. An example IPT consisting of geometrically modified images is presented in Figure~\ref{Fig:Geom_IPT}. Note that the IPT configuration is the same as the one used to evaluate photometrically modified images. We have conducted the experiment using the following two protocols, and evaluated the performance using root identification accuracy at Ranks 1, 2 and 3 and IPT reconstruction accuracy.

 1. The first protocol involves training on the photometrically modified images, while testing on geometrically modified images. In this protocol, the training set did not include any geometrically modified images, so it assesses the robustness of the basis functions on different classes of transformations (\textit{i.e.}, photometric versus geometric). We have \textit{not} modified the asymmetric measure computation method or the tree-spanning method used in constructing the IPT.  
 
 2.  The second protocol involves training and testing on geometrically modified images. To accomplish this task, we created a new training set of 5,865 pairs of original and geometrically transformed images using the LFW dataset. The objective is to evaluate the performance of the basis functions when trained and tested on geometric modifications, unlike in the first protocol.  

We compared the performance of the proposed method with a baseline algorithm described in~\cite{Dias_12_MST}. The baseline algorithm uses Speeded-Up Robust Features (SURF) and RANSAC algorithm for the task of geometric registration, followed by color channel normalization, and uses the residual photometric error as the asymmetric measure. The Oriented Kruskal algorithm is used for spanning the IPT which is a minimal spanning tree in their case. We implemented the baseline algorithm in two ways. 

    (a) Firstly, we used SURF and M-SAC (M-estimator sample and consensus scheme which is an improved variant of RANSAC) for geometric registration. We did not perform color channel normalization since we used gray-scale images. We then rescaled pixel intensities in the original and modified images to $[0,255]$ prior to evaluation. We used the Oriented Kruskal algorithm for constructing the IPT.
    
    (b) Secondly, we used the best performing basis function to compute the likelihood ratio to be used as the asymmetric measure and then employed the Oriented Kruskal for spanning the IPT.

It is important to note that, unlike the baseline method, the proposed method does not require any separate geometric registration for modeling the geometric transformations.

\begin{table}[]
\centering
\caption{Experiment 5: Geometric transformations and their parameter ranges used in this work.}
\label{Tab:GeomParams}
\begin{tabular}{|ll|l|}
\hline
\multicolumn{2}{|l|}{\textbf{Geometric transformations}} & \textbf{Parameters}    \\ \hline \hline
\multicolumn{2}{|l|}{Re-sampling}               & {[}90\%, 110\%{]} \\ \hline
\multirow{3}{*}{Generic Affine} & Rotation    & {[}-5$^{\circ}$, 5$^{\circ}${]}   \\
                                & Translation & {[}5, 20{]}   \\
                                & Scaling     & {[}90\%, 110\%{]} \\ \hline
\end{tabular}
\end{table}

\subsubsection{Experiment 6: Ability to handle near-duplicates available online}
Near-duplicate images of celebrities and political figures are widely circulated on the internet. The actual sequence of generation of such near-duplicates may be unknown, but these images represent pragmatic scenarios where the ground truth may not be always available. In this experiment, we analyze how the proposed asymmetric measure and IPT construction methods can handle such images. To this end, we followed the suggestion presented in~\cite{NDDR} and used Google image search to download 40 near-duplicates retrieved from the following 5 queries: \textit{Angelina Jolie, Kate Winslet, Superman, Britney Spears} and \textit{Bob Marley}. We used training parameter distributions learnt from both geometrically and photometrically modified images for each of the five basis functions used in this work. We then used the proposed asymmetric measure computation method to identify top 3 candidate root nodes. For each of the candidate root node we then reconstructed an IPT. Due to unavailability of ground truth, we could not evaluate the accuracy of the reconstructed IPTs, but we present qualitative assessment of the reconstructed IPTs.

\subsubsection{Experiment 7: Ability to handle deep learning-based transformations and image augmentation schemes}
Several deep learning-based transformations and image augmentation packages are available that can be used for applying sophisticated transformations to images in an automated fashion generating a large number of near-duplicates. In this experiment, we used images generated using a deep learning-based autoencoder~\cite{CAE} and open source image augmentation packages~\cite{Augmentor}. We conducted the experiment using two protocols.

\begin{figure}
\centering
\subfloat[]
{
    \includegraphics[scale=.33]{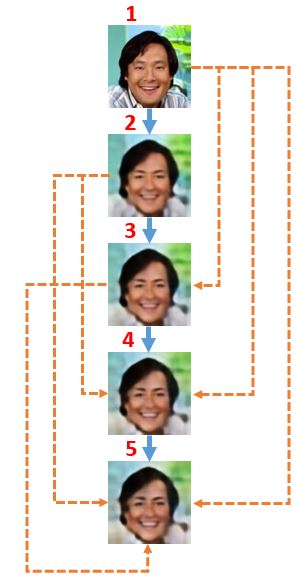} 
} \hspace{0.7cm}
\subfloat[]
{
    \includegraphics[scale=.33]{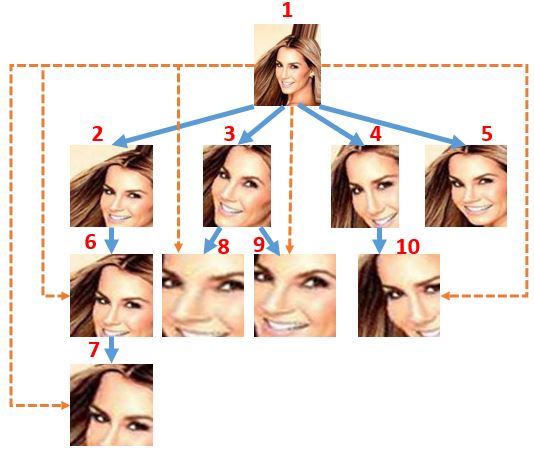} 
}
\caption{Experiment 7: (Left) IPT test configuration used for evaluating proposed method on autoencoder generated near-duplicates. (Right) IPT test configuration used for evaluating proposed method on images generated by open source image augmentation package. The bold arrows indicate immediate links and the dashed arrows indicate ancestral links.}
\label{Fig: DL}
\end{figure}

1. The first protocol involves a deep convolutional autoencoder~\cite{CAE}. The autoencoder was trained on $\sim19,000$ images from the CelebA dataset~\cite{CelebA} to generate 80 near-duplicate images belonging to 16 subjects. The resultant IPT configuration is depicted in Figure~\ref{Fig: DL}(a). The convolutional autoencoder comprises of an encoder block that consists of five convolutional layers, followed by ReLU after each convolutional layer, and the decoder block comprises of traditional convolutional layers and nearest-neighbor based upsampling.\footnote{\url{https://sebastianraschka.com/deep-learning-resources.html}} We did not use de-convolution or transposed convolution layers, as they can lead to checkerboard artifacts. The intuition behind using an autoencoder for generating near-duplicate images is to leverage its ability to perform high fidelity reconstruction of the original input images. This fits the definition of `near-duplicates' in our image phylogeny task and has, therefore, been used in this experiment. We apply the original image as an input to the autoencoder to generate the first set of near-duplicates at depth=1. This first set of reconstructed images are again fed as input to the same autoencoder to generate near-duplicates at depth=2, and so on until we generate near-duplicates at depth=5. 
    
 2. The second protocol involves Augmentor~\cite{Augmentor}, a data augmentation tool used when training deep neural networks. We used this tool, which is an open source package in Python, to apply random distortions such as zoom, cropping, rotation, re-sampling and elastic deformations on an image. See Figure~\ref{Fig: DL}(b). Some of these image transformations and their diverse parameter ranges (training involved rotation values in the interval $[-5^{\circ}, 5^{\circ}]$, whereas, testing using Augmentor involved rotation values in the interval $[-10^{\circ}, 10^{\circ}]$) are not encountered during the training stage. We randomly selected 100 images belonging to 100 subjects from the CelebA dataset. We applied the Augmentor on each of these 100 images to create 100 IPTs. Each IPT contains 10 images. So we tested on a total of 1,000 near-duplicate images.  
    
\begin{figure}
\centering
\subfloat[]
{
    \includegraphics[scale=.36]{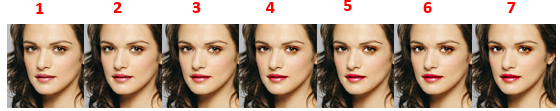} 
}
\subfloat[]
{
    \includegraphics[scale=.36]{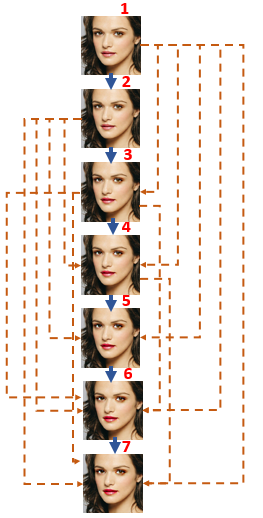} 
}
\caption{Experiment 7: (Left) Near-duplicates generated using BeautyGlow generative network. (Right) IPT constructed using Chebsyshev polynomials for the near-duplicates on the left. The bold arrows indicate immediate links and the dashed arrows indicate ancestral links.}
\label{Fig: BG}
\end{figure}

In addition, we also used some images synthesized using a deep learning-based generative network known as BeautyGlow~\cite{BeautyGlow}. The generative network performs a style transfer on the makeup of the individual in face images, resulting in near-duplicates as shown in Figure~\ref{Fig: BG}(a). Images are generated by sequentially increasing the magnification value of the makeup, highlighting the intensity of the makeup. We used 13 IPTs (each IPT contains 7 images), resulting in a total of 91 images. 

\section{Results and Analysis}
\label{Sec:Results}
In this section, we first report the results observed for the experiments described in the previous section, and we further present our insights into the findings.

\subsection{Results of Experiment 1}

\begin{figure*}

\subfloat[Legendre]{ 
\includegraphics[scale=.24]{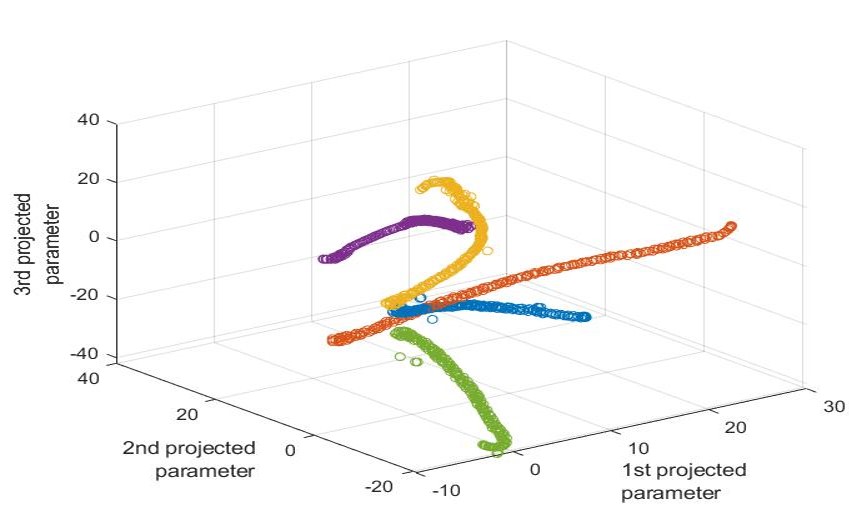} \hspace{-3.8cm}\raisebox{\dimexpr 3.1cm+\height}{\footnotesize{\textcolor{red}{Brightness transformation}}}  \hspace{1cm}
\includegraphics[scale=.24]{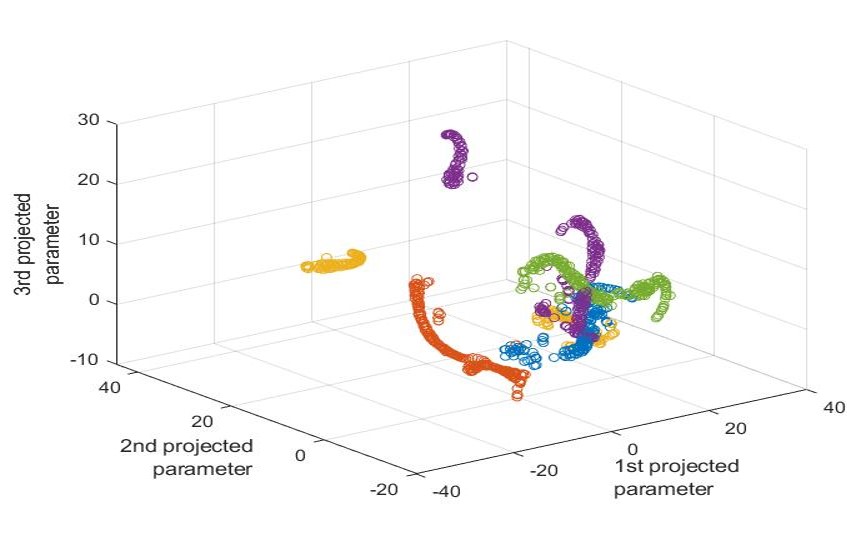} \hspace{-3.2cm}\raisebox{\dimexpr 3.1cm+\height}{\footnotesize{\textcolor{red}{Gamma adjustment}}} \hspace{1cm}  \\ 
\includegraphics[scale=.24]{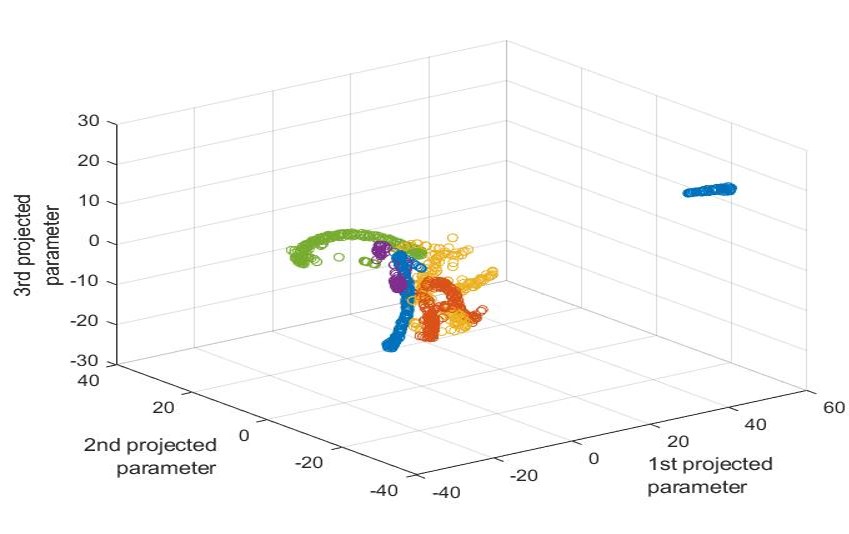} \hspace{-3.2cm}\raisebox{\dimexpr 3.1cm+\height}{\footnotesize{\textcolor{red}{Gaussian smoothing}}} \hspace{0.8cm}
\includegraphics[scale=.24]{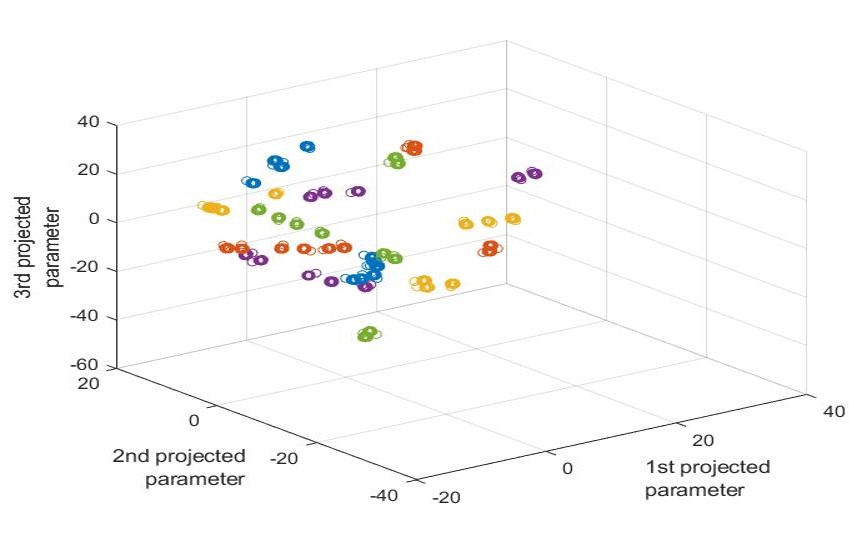} \hspace{-3.1cm}\raisebox{\dimexpr 3.1cm+\height}{\footnotesize{\textcolor{red}{Median filtering}}} \hspace{1cm}
}

\subfloat[Chebyshev]{ 
\includegraphics[scale=.24]{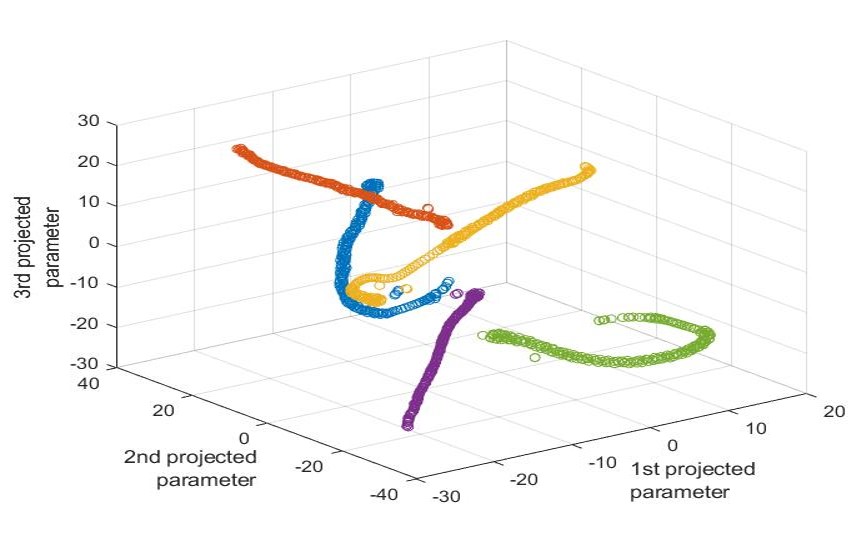}  \hspace{0.08cm} 
\includegraphics[scale=.24]{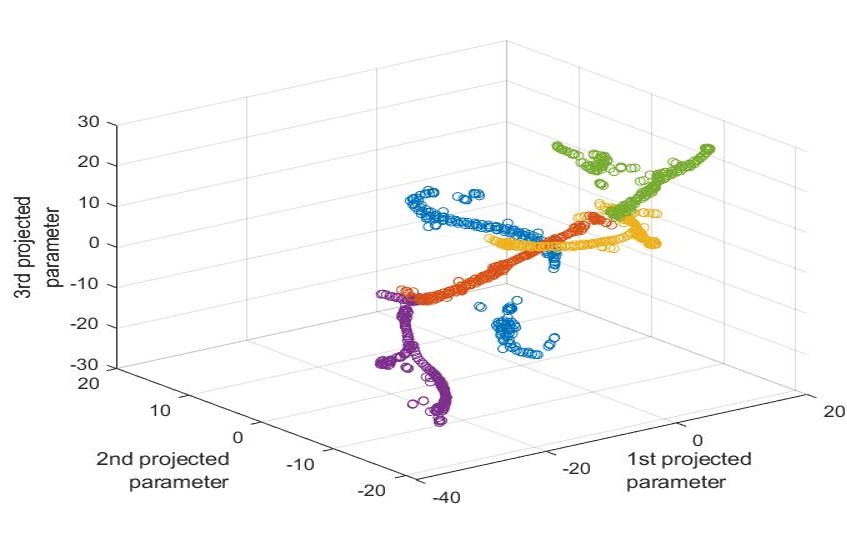}  \hspace{0.03cm} \\
\includegraphics[scale=.24]{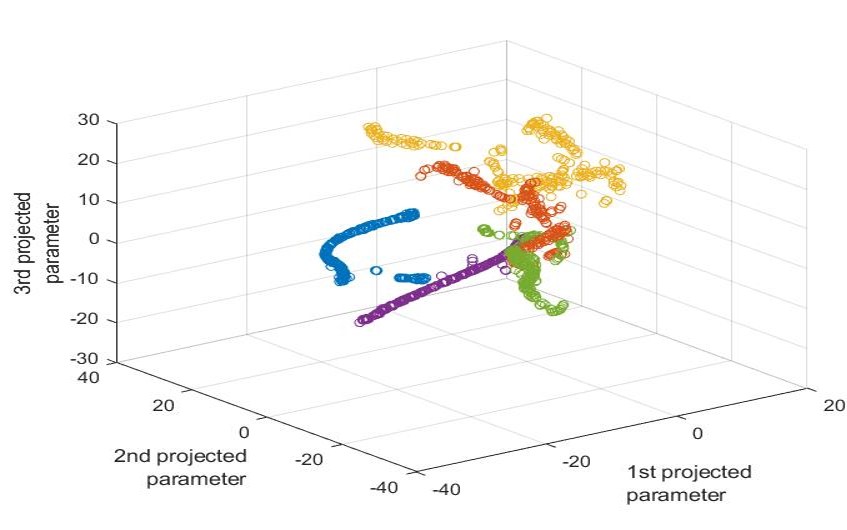} \hspace{0.03cm} 
\includegraphics[scale=.24]{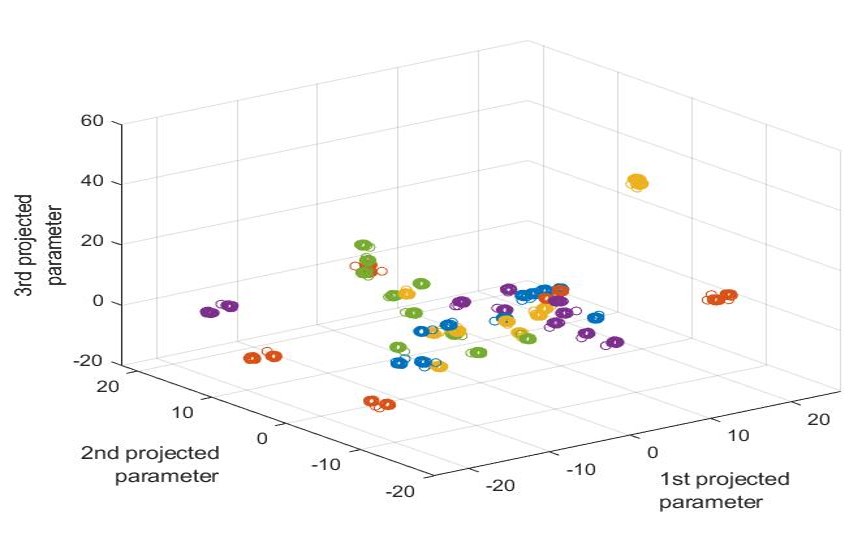} \hspace{0.08cm}
}

\subfloat[Gabor]{ 
\includegraphics[scale=.24]{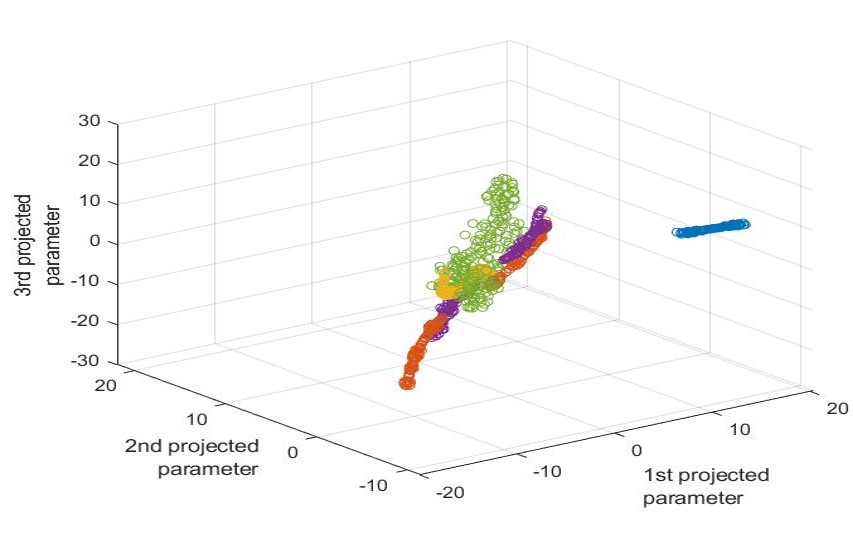}  \hspace{0.08cm} 
\includegraphics[scale=.24]{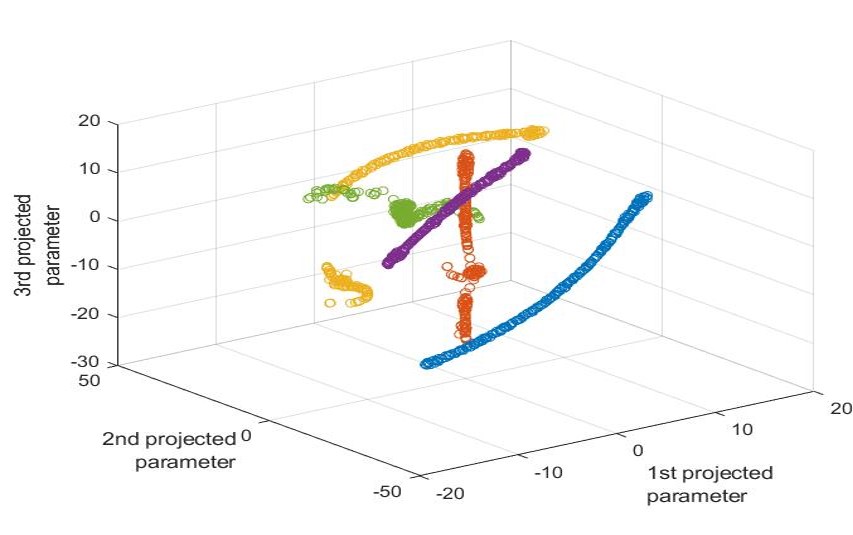}  \hspace{0.03cm} \\
\includegraphics[scale=.24]{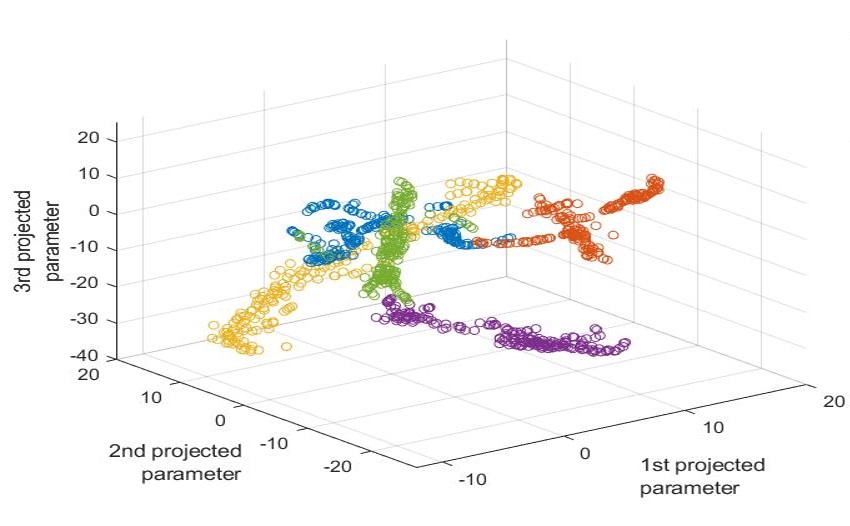}  \hspace{0.03cm}
\includegraphics[scale=.24]{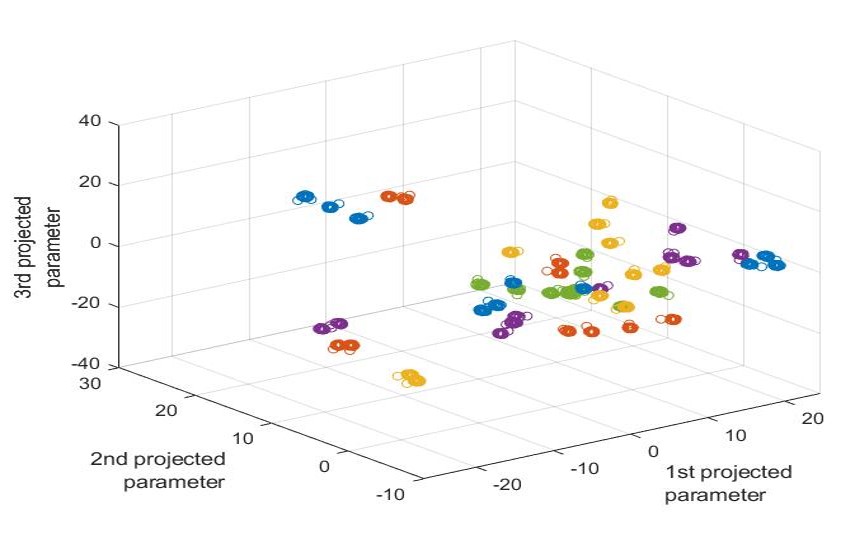}  \hspace{0.08cm}
}

\subfloat[Gaussian RBF]{ 
\includegraphics[scale=.24]{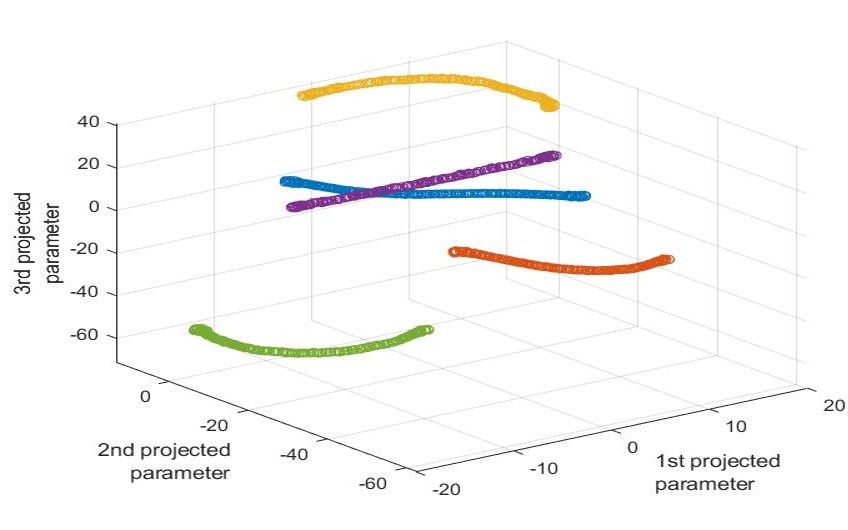} \hspace{0.08cm}
\includegraphics[scale=.24]{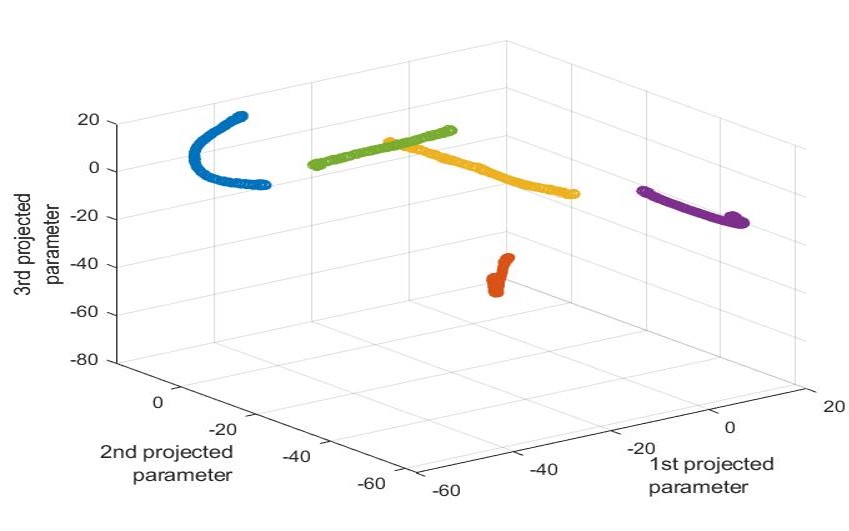}    \hspace{0.03cm} \\
\includegraphics[scale=.24]{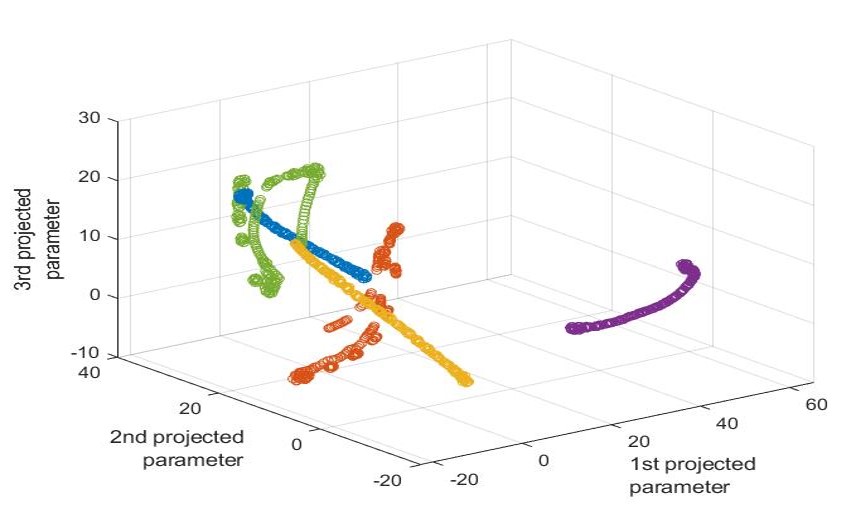}  \hspace{0.03cm}
\includegraphics[scale=.24]{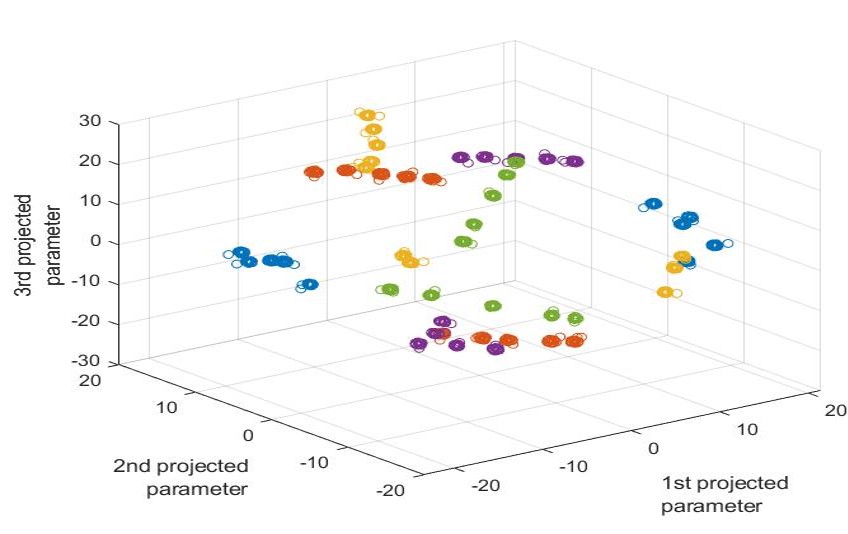}  \hspace{0.08cm}
}

\subfloat[Bump RBF]{ 
\includegraphics[scale=.24]{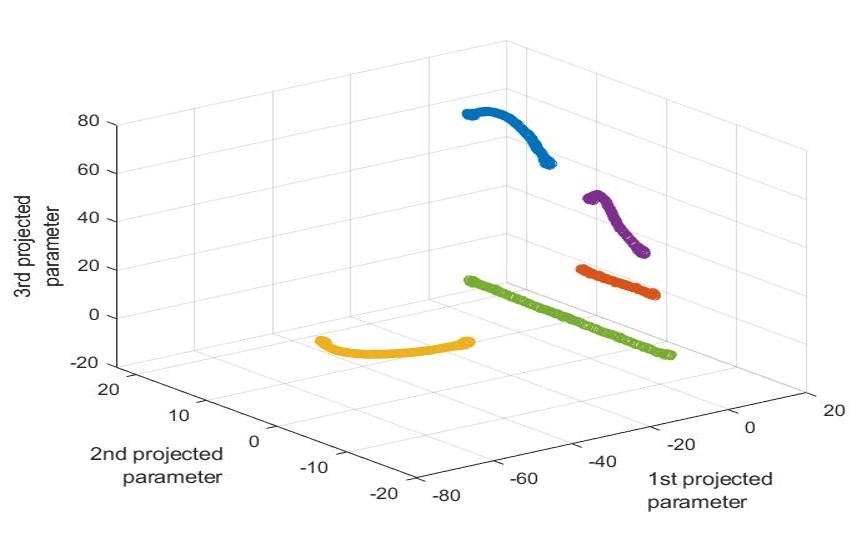}  \hspace{0.08cm}
\includegraphics[scale=.24]{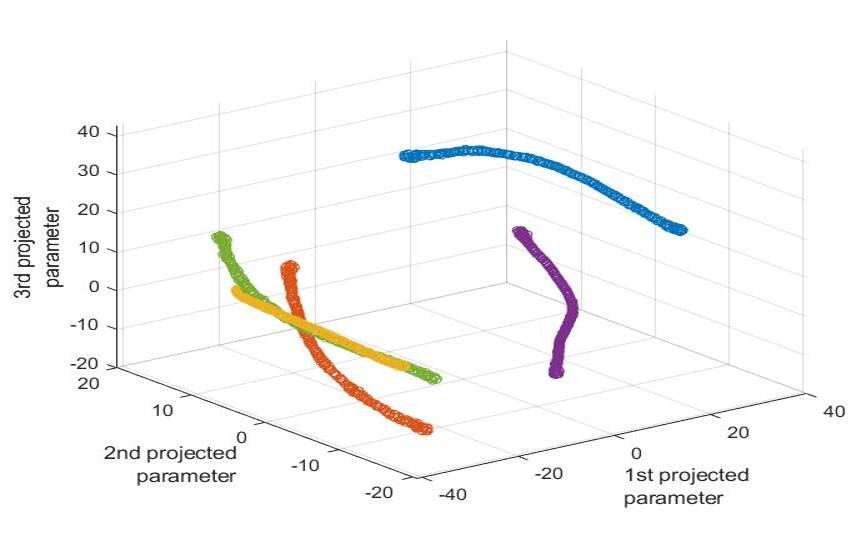}    \hspace{0.03cm} \\
\includegraphics[scale=.24]{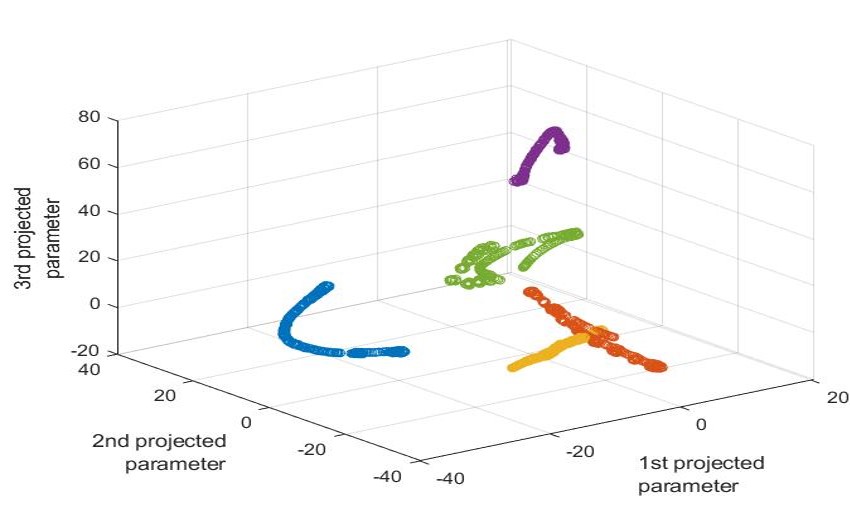}   \hspace{0.03cm}
\includegraphics[scale=.24]{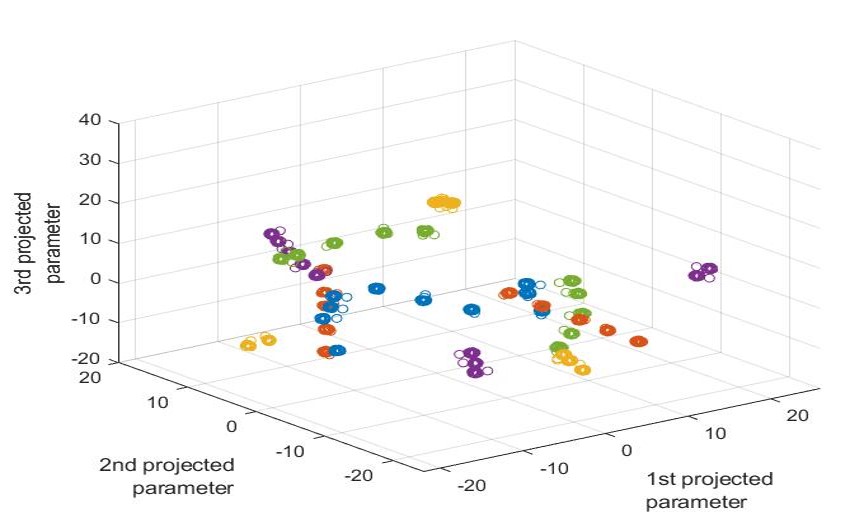}  \hspace{0.08cm}
}

\hspace{6cm} \includegraphics[scale=0.35]{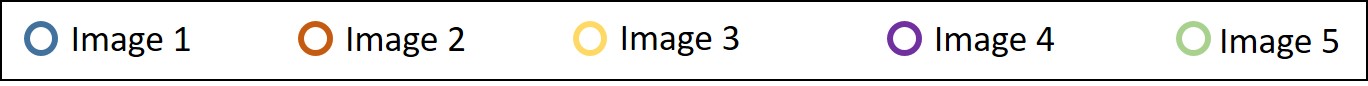}
\caption{Experiment 1: 3D projected parameters using $t$-SNE corresponding to each photometric transformation (column) modeled using each basis function (row). Each color represents a single image. A total of 5 images were modeled. Gaussian and Bump RBFs model majority of the transformations reasonably well as indicated by the last two rows. The Brightness transformation was easiest to model as the parameters of the basis functions follow a continuous path.}
\label{Fig: Model}
\end{figure*}

\begin{figure}
\centering
\includegraphics[scale=0.25]{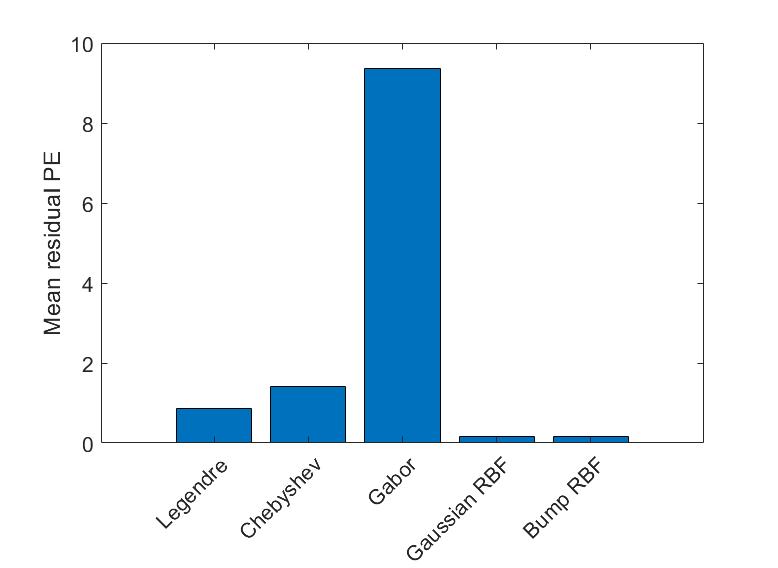}
\caption{Experiment 1: The photometric error between the actual output and the the output modeled using the basis functions is denoted as residual photometric error (PE). The mean of the residual PE is demonstrated for 2,000 image pairs modeled in both forward and reverse directions using the five basis functions. Gabor resulted in the highest residual PE, and the RBFs yield the lowest residual PE demonstrating their efficacy in reliably modeling the transformations.}
\label{Fig: Boxplot}
\end{figure}

The 3D projected vectors obtained using $t$-SNE are illustrated in Figure~\ref{Fig: Model} for each of the transformations modeled using the five basis functions. Each column denotes a photometric transformation, and each row denotes a basis function. As evident from the projections, the basis functions can model the majority of the transformations fairly well. The parameters governing each transformation are incrementally modified and, hence, their projections should ideally span a continuous trajectory. Also, we expect to observe this behavior irrespective of the transformations used or the identity of the subject. We indeed observe such a behavior for most of the cases, except for Gamma adjustment, where the polynomials and wavelet functions flounder. Note that median filtering requires integer parameter values (height and width of window). Therefore, in the $t$-SNE results (last column in Figure~\ref{Fig: Model}) we observe small clusters, depicting accurate modeling of discrete parameterized transformations. Out of all the basis functions, the radial basis functions seem to model the transformations the best. Figure~\ref{Fig: Boxplot} further substantiates that the RBFs are best at modeling transformations while Gabor wavelets perform relatively poorly. The RBFs result in the lowest mean residual photometric error, suggesting a more accurate modeling of the photometric transformations.

In terms of discriminability between the forward and reverse directions, the projections are almost indistinguishable in the two directions for the wavelet functions, but they are relatively better for polynomial functions and the RBFs, as evidenced in Figure~\ref{Fig:For_Rev_Modeling}. The polynomials have fairly well-separated projections, indicating their ability to discriminate between the forward and reverse directions. We anticipate that this ability will be reflected in the IPT reconstruction experiments. 

\subsection{Results of Experiment 2}
\begin{figure*}[h]
\centering
\subfloat[]
{
    \includegraphics[scale=.11]{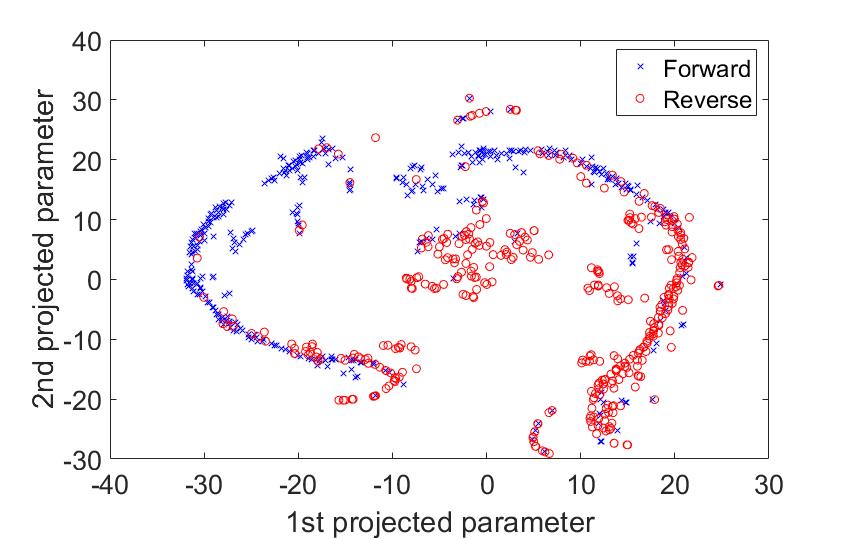} 
}\hspace{0.1cm}
\subfloat[]
{
    \includegraphics[scale=.11]{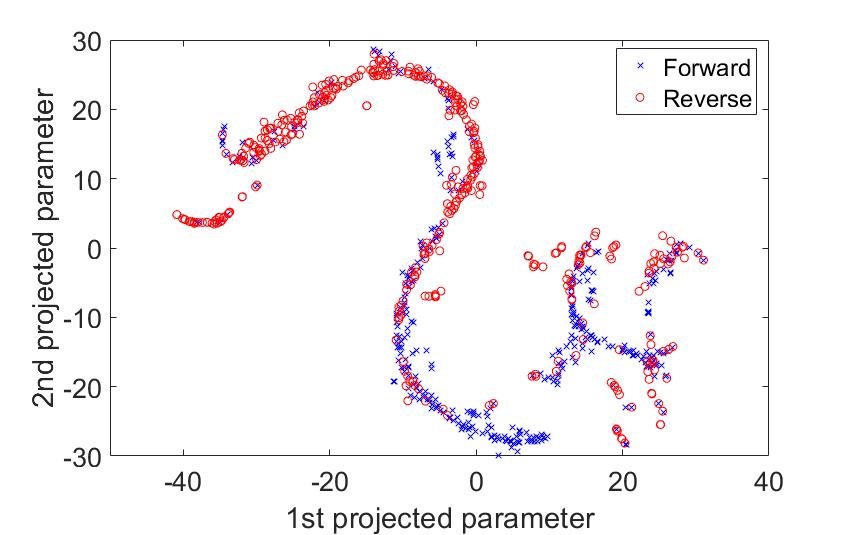} 
}\hspace{0.1cm}
\subfloat[]
{
    \includegraphics[scale=.11]{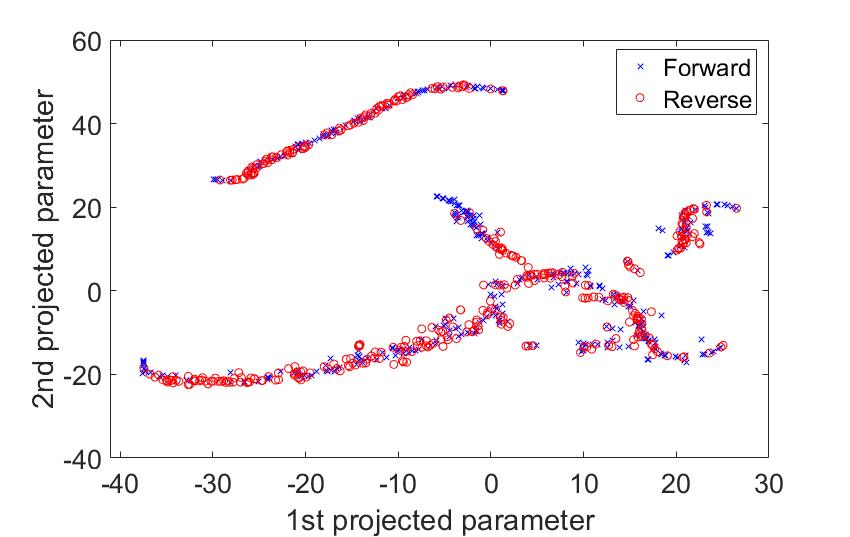} 
}\hspace{0.1cm}
\subfloat[]
{
    \includegraphics[scale=.11]{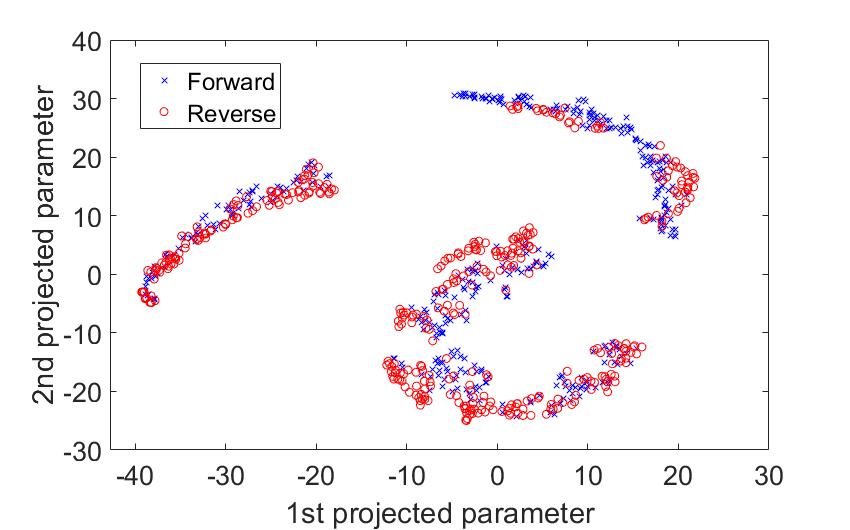} 
}\hspace{0.1cm}
\subfloat[]
{
    \includegraphics[scale=.11]{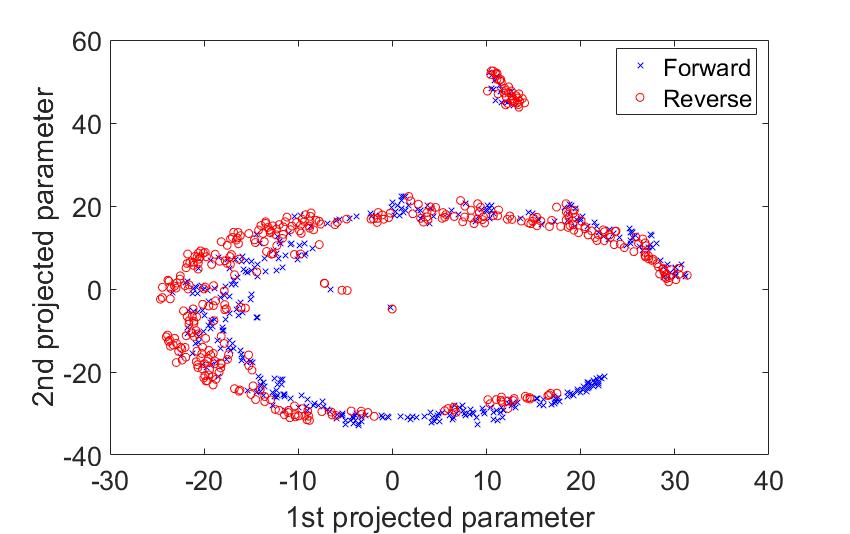} 
}

\caption{Experiment 1: 2D projected parameters using $t$-SNE in forward and reverse directions, corresponding to all 4 transformations modeled using each basis function: (a) Legendre, (b) Chebyshev, (c) Gabor, (d) Gaussian RBF and (e) Bump RBF. Legendre and Chebyshev polynomials can better discriminate between forward and reverse directions as indicated by the relatively better separated parameter distributions compared to the remaining basis functions.}
\label{Fig:For_Rev_Modeling}
\end{figure*}

The results of root identification and IPT reconstruction are presented in Tables~\ref{Tab:Face_Set1} and \ref{Tab:Face_Set2}. Results indicate that polynomials (Legendre and Chebyshev) perform the best in a majority of cases among the set of five basis functions selected in this work. The results are consistent with the observations reported in Figure~\ref{Fig:For_Rev_Modeling}, which indicates sufficient discriminability offered by the polynomials. For the partial set, Legendre polynomials perform best both in terms of root identification (89.91\%) and IPT reconstruction (70.61\%) accuracies, closely followed by Chebyshev polynomials. For the full set, Gaussian RBF performs the best in terms of root identification accuracy (80.85\%) while Chebsyhev polynomials perform the best in terms of IPT reconstruction accuracy (66.54\%, a small improvement of $\approx$1.5\% is observed compared to the results in~\cite{FacePhyloTree_2019}). 

\subsection{Results of Experiment 3}
The results for the cross-modality experiments are presented in Table~\ref{Tab:Iris} for iris images, and in Tables~\ref{Tab:Fingerprint_Config1} and~\ref{Tab:Fingerprint_Config2} for fingerprint images. The purpose of this set of experiments is to assess how the proposed method performs for (i) the same IPT configuration but used across two different modalities, and (ii) the same modality but tested on different IPT configurations. Note that the training modality is different than the test modality in both cases. 

The results in Table~\ref{Tab:Iris} indicate that Chebyshev polynomials obtain 94.90\% root identification accuracy at Rank 3 and an IPT reconstruction accuracy of 67.90\% for iris images. It is closely followed by Legendre polynomials. However, other basis functions perform poorly, specifically, the Gabor wavelets. Gabor wavelets are good texture descriptors, \ie they can extract the high-frequency features reliably from an image. In the case of photometric transformations, the pixel intensity gradient, which contributes toward high-frequency features are not significantly affected and, thus, the wavelet fails to correctly model the transformation between an image pair.

Image phylogeny on near-duplicate fingerprint images is extremely hard, as evident from the results in Tables~\ref{Tab:Fingerprint_Config1} and~\ref{Tab:Fingerprint_Config2}. Visual inspection reveals that the set of near-duplicate fingerprint images appear to be black blobs on a white background, with no discernible differences between the set. Therefore, the root identification and IPT reconstruction accuracies are worse compared to the face and iris modalities: the best root identification performance is 65.28\% and IPT reconstruction accuracy is 70.59\%. This experiment also shows that the performance varies across configurations. Specifically, symmetric configurations (Config-II) can be more difficult to reconstruct than asymmetric configurations (Config-I).

Next, we compare the results of the cross-modality experiments with the baseline, which is the intra-modality experiment described in Section~\ref{subsubsect:expt3}. The results of the baseline experiments are reported in Table~\ref{Tab:IntraModality}. The results indicate that cross-modality performance is commensurate with the intra-modality performance. For example, for iris images, the intra-modality experiment obtains the best root identification accuracy of 94.63\% at Rank 3, while the cross-modality experiment obtains the best root identification accuracy of 94.90\% at Rank 3. Furthermore, the intra-modality experiment obtains the highest IPT reconstruction accuracy of 68.62\%, while the cross-modality experiment obtains the highest IPT reconstruction accuracy of 67.90\% in the case of iris images. 

\subsection{Results of Experiment 4}
The training set comprised of images modified using 4 rudimentary photometric transformations. In the real world, a plethora of image editing applications exist, thereby making image phylogeny for face images a generally difficult problem. We hypothesize that by creating a training dataset through random parameters on simple transformations, the unseen transformations can be reliably modeled. Results reported in Table~\ref{Tab:Photoshop} indicate that unseen transformations were modeled fairly well. Legendre polynomials performed the best in terms of root identification accuracy with 76.47\% averaged across the four IPT configurations (see Figure~\ref{Fig:Photoshop}). Chebyshev polynomials performed the best in terms of IPT reconstruction accuracy with 76.25\% averaged across the four IPT configurations (a small improvement of $\approx$1.67\% is observed compared to the results in~\cite{FacePhyloTree_2019}).

\begin{table}[]
\centering
\caption{Experiment 2: Root identification and IPT reconstruction accuracies for face images (Partial Set). }
\label{Tab:Face_Set1}
\scalebox{0.98}{
\begin{tabular}{|l|ll|} \hline
Basis Function & \begin{tabular}[c]{@{}l@{}}Root identification (\%)\\ Rank 1/2/3\end{tabular} & \begin{tabular}[c]{@{}l@{}}IPT Reconstruction\\ (\%)\end{tabular} \\ \hline \hline
Legendre       & \textbf{65.90/82.18/89.91}                                                          & \textbf{70.61                                                          } \\
Chebyshev      & 53.62/74.69/85.52                                                           & 70.53                                                           \\
Gabor          & 27.66/41.74/56.06                                                           & 55.54                                                           \\
Gaussian RBF   & 65.25/81.04/87.79                                                           & 66.15                                                           \\
Bump RBF       & 63.79/80.07/86.41                                                           & 66.52                                                          \\ \hline
\end{tabular}}
\end{table}

\begin{table}[]
\centering
\caption{Experiment 2: Root identification and IPT reconstruction accuracies for face images (Full set). }
\label{Tab:Face_Set2}
\scalebox{0.98}{
\begin{tabular}{|l|ll|} \hline
Basis Function & \begin{tabular}[c]{@{}l@{}}Root identification (\%)\\ Rank 1/2/3\end{tabular} & \begin{tabular}[c]{@{}l@{}}IPT Reconstruction\\ (\%)\end{tabular} \\ \hline \hline
Legendre       & 50.45/66.74/75.68                                                         & 65.05   \\
Chebyshev      & 45.18/65.13/76.86                                                         & \textbf{66.54} \\
Gabor          & 29.48/44.77/58.01                                                           & 55.46 \\
Gaussian RBF   & \textbf{56.44/71.87/80.85}                                                          & 63.84 \\
Bump RBF       & 55.34/70.85/80.09                                                           & 64.27 \\ \hline
\end{tabular}}
\end{table}

\begin{table}[]
\centering
\caption{Experiment 3A: Root identification and IPT reconstruction accuracies for iris images in the cross-modality setting. }
\label{Tab:Iris}
\scalebox{0.98}{
\begin{tabular}{|l|ll|} \hline
Basis Function & \begin{tabular}[c]{@{}l@{}}Root identification (\%)\\ Rank 1/2/3\end{tabular} & \begin{tabular}[c]{@{}l@{}}IPT Reconstruction\\ (\%)\end{tabular} \\ \hline \hline
Legendre       & 56.75/76.58/87.88                                                       & 67.53  \\
Chebyshev      & \textbf{72.59/88.29/94.90}                                                        & \textbf{67.90} \\
Gabor          & 5.79/12.40/19.70                                                           & 51.23 \\
Gaussian RBF   & 40.08/63.64/76.45                                                         & 66.74 \\
Bump RBF       & 39.67/60.88/76.03                                                          & 66\\ \hline
\end{tabular}}
\end{table}

\begin{table}[]
\centering
\caption{Experiment 3B: Root identification and IPT reconstruction accuracies for fingerprint images (Config -I) in the cross-modality setting.}
\label{Tab:Fingerprint_Config1}
\scalebox{0.98}{
\begin{tabular}{|l|ll|} \hline
Basis Function & \begin{tabular}[c]{@{}l@{}}Root identification (\%)\\ Rank 1/2/3\end{tabular} & \begin{tabular}[c]{@{}l@{}}IPT Reconstruction\\ (\%)\end{tabular} \\ \hline \hline
Legendre       & 29.66/44.32/56.82                                                       & 68.99  \\
Chebyshev      & \textbf{31.93}/46.14/57.39                                                         & \textbf{70.59} \\
Gabor          & 22.50/37.27/51.36                                                          & 68.08 \\
Gaussian RBF   & 30.80/50.34/61.14                                                        & 68.98 \\
Bump RBF       & 31.59/\textbf{51.70/62.50 }                                                         & 68.51\\ \hline
\end{tabular}}
\end{table}

\begin{table}[]
\centering
\caption{Experiment 3B: Root identification and IPT reconstruction accuracies for fingerprint images (Config -II) in the cross-modality setting.}
\label{Tab:Fingerprint_Config2}
\scalebox{0.98}{
\begin{tabular}{|l|ll|} \hline
Basis Function & \begin{tabular}[c]{@{}l@{}}Root identification (\%)\\ Rank 1/2/3\end{tabular} & \begin{tabular}[c]{@{}l@{}}IPT Reconstruction\\ (\%)\end{tabular} \\ \hline \hline
Legendre       & 34.58/51.11/59.31                                                       & 65.82  \\
Chebyshev      & \textbf{35/55.28/65.28}                                                         & \textbf{65.93} \\
Gabor          & 31.39/48.75/63.33                                                         & 60.76 \\
Gaussian RBF   & 14.58/23.75/36.81                                                        & 59.29 \\
Bump RBF       & 17.50/28.33/40.42                                                         & 59.96\\ \hline
\end{tabular}}
\end{table}

\begin{table}[]
\centering
\caption{Experiment 3: Baseline performance of basis functions in terms of root identification and IPT reconstruction accuracies in the intra-modality setting.}
\label{Tab:IntraModality}
\scalebox{0.88}{
\begin{tabular}{|l|l|ll|} \hline
\begin{tabular}[c]{@{}l@{}}Modality \& \\ Configuration\end{tabular}                & \begin{tabular}[c]{@{}l@{}}Basis \\ Function\end{tabular} & \begin{tabular}[c]{@{}l@{}}Root identification (\%)\\ Rank 1/2/3\end{tabular} & \begin{tabular}[c]{@{}l@{}}IPT\\ Reconstruction (\%)\end{tabular} \\ \hline
\multirow{5}{*}{IRIS}                                                               & Legendre                                                  & 64.46/83.75/91.46                                                           & \textbf{68.62}                                                  \\
                                                                                    & Chebyshev                                                 & \textbf{76.31/89.53/94.63}                                                  & 66.62                                                           \\
                                                                                    & Gabor                                                     & 8.54/18.60/27.55                                                            & 54.33                                                           \\
                                                                                    & Gaussian RBF                                              & 29.89/49.31/66.25                                                           & 60.77                                                           \\
                                                                                    & Bump RBF                                                  & 25.21/38.15/48.76                                                           & 58.06                                                           \\ \hline \hline
\multirow{5}{*}{\begin{tabular}[c]{@{}l@{}}FINGERPRINT\\ \\ Config-I\end{tabular}}  & Legendre                                                  & 34.69/48.13/57.81                                                           & \textbf{71.92}                                                  \\
                                                                                    & Chebyshev                                                 & \textbf{38.75/58.44/67.55}                                                  & 71.46                                                           \\
                                                                                    & Gabor                                                     & 5.94/14.06/23.75                                                            & 64.37                                                           \\
                                                                                    & Gaussian RBF                                              & 28.13/48.13/59.69                                                           & 66.96                                                           \\
                                                                                    & Bump RBF                                                  & 39.06/56.56/66.56                                                           & 68.35                                                           \\ \cline{2-4}
\multirow{5}{*}{\begin{tabular}[c]{@{}l@{}}FINGERPRINT\\ \\ Config-II\end{tabular}} & Legendre                                                  & 35.63/50.31/63.12                                                           & \textbf{66.22}                                                  \\
                                                                                    & Chebyshev                                                 & \textbf{42.50/56.56/69.69}                                                  & 65.52                                                           \\
                                                                                    & Gabor                                                     & 5.31/9.69/17.81                                                             & 55.23                                                           \\
                                                                                    & Gaussian RBF                                              & 26.56/42.19/54.37                                                           & 59.91                                                           \\
                                                                                    & Bump RBF                                                  & 31.56/48.44/60.31                                                           & 63.44                                                          \\ \hline
\end{tabular}}
\end{table}

\begin{table*}[]
\centering
\caption{Experiment 4: Root identification and IPT reconstruction accuracies for unseen photometric transformations.}
\label{Tab:Photoshop}
\scalebox{0.8}{
\begin{tabular}{|l|ll|ll|ll|ll|} \hline
\multirow{3}{*}{Basis Functions} & \multicolumn{8}{c|}{IPT Configurations}                                                                                                                                                                                                                                                                                                                                                                                                                                                                                                                                                                                                                                                                                                          \\
                                 & \multicolumn{2}{c}{IPT 1}                                                                                                                                                          & \multicolumn{2}{c}{IPT 2}                                                                                                                                                          & \multicolumn{2}{c}{IPT 3}                                                                                                                                                         & \multicolumn{2}{c|}{IPT 4}                                                                                                                                                         \\ \cline{2-9}
                                 & \multicolumn{1}{l}{\begin{tabular}[l]{@{}l@{}}Root identification\\ Rank 3 (\%)\end{tabular}} & \multicolumn{1}{l|}{\begin{tabular}[l]{@{}l@{}}IPT \\ Reconstruction (\%)\end{tabular}} & \multicolumn{1}{l}{\begin{tabular}[l]{@{}l@{}}Root identification\\ Rank 3 (\%)\end{tabular}} & \multicolumn{1}{l|}{\begin{tabular}[l]{@{}l@{}}IPT \\ Reconstruction (\%)\end{tabular}} & \multicolumn{1}{l}{\begin{tabular}[l]{@{}l@{}}Root identification\\ Rank 3 (\%)\end{tabular}} & \multicolumn{1}{l|}{\begin{tabular}[l]{@{}l@{}}IPT\\ Reconstruction (\%)\end{tabular}} & \multicolumn{1}{l}{\begin{tabular}[l]{@{}l@{}}Root identification\\ Rank 3 (\%)\end{tabular}} & \multicolumn{1}{l|}{\begin{tabular}[l]{@{}l@{}}IPT\\ Reconstruction (\%)\end{tabular}} \\ \cline{2-9} \cline{2-9}
Legendre                         & \textbf{66.67}                                                                              & 82.22                                                                                & \textbf{90}                                                                                 & 71.67                                                                                & \textbf{77.78}                                                                              & 44.44                                                                               & \textbf{71.43}                                                                              & \textbf{100}                                                                        \\
Chebyshev                        & 44.44                                                                                       & \textbf{84.44}                                                                       & 60                                                                                          & \textbf{75}                                                                          & 44.44                                                                                       & 45.56                                                                               & \textbf{71.43}                                                                              & \textbf{100}                                                                        \\
Gabor                            & 44.44                                                                                       & 82.22                                                                                & 80                                                                                          & 71.67                                                                                & \textbf{77.78}                                                                              & \textbf{48.89}                                                                      & \textbf{71.43}                                                                              & \textbf{100}                                                                        \\
Gaussian RBF                     & \textbf{66.67}                                                                              & \textbf{84.44}                                                                       & 40                                                                                          & 68.33                                                                                & 44.44                                                                                       & 43.33                                                                               & \textbf{71.43}                                                                              & \textbf{100}                                                                        \\
Bump RBF                         & \textbf{66.67}                                                                              & \textbf{84.44}                                                                       & 60                                                                                          & 68.33                                                                                & 33.33                                                                                       & 44.44                                                                               & \textbf{71.43}                                                                              & \textbf{100} \\ \hline                                                                      
\end{tabular}}
\end{table*}

\begin{figure}[h]
\centering
\subfloat[]
{
    \includegraphics[scale=.38]{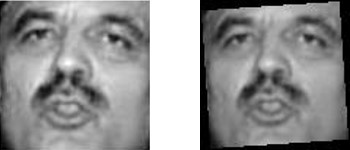} 
} \hfill
\subfloat[]
{
    \includegraphics[scale=.38]{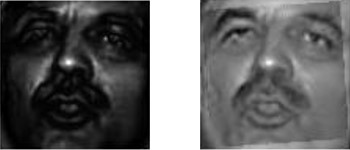} 
}\hfill
\subfloat[]
{
    \includegraphics[scale=.38]{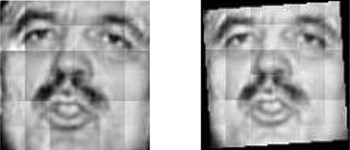} 
}\hfill

\caption{Experiment 5: Example of geometric transformation (rotation) modeling using basis functions. (a) Original image (on the left) and the transformed image (on the right). (b) Modeled image pair using Legendre polynomials (modeled original image is on the left and modeled transformed image is on the right). (c) Modeled image pair using Gaussian RBF (modeled original image is on the left and modeled transformed image is on the right).}
\label{Fig:Geom_modeling}
\end{figure}

\subsection{Results of Experiment 5}

We present examples of geometric transformation modeling using the basis functions in Figure~\ref{Fig:Geom_modeling}. We observe that Gaussian RBFs outperform Legendre polynomials at modeling the geometric transformations (see Figure~\ref{Fig:Geom_modeling}) due to two reasons - (i) the radial basis functions can potentially span infinite range of values as opposed to the polynomials which can span values within a finite interval, and (ii) the RBFs did patch-level modeling compared to the pixel-level modeling done by the polynomials. The results in Table~\ref{Tab:Geom} indicate that the basis functions perform significantly better when trained on geometrically modified images (second protocol) compared to when trained on photometrically modified images (first protocol). As anticipated, if the class of transformations are the same in both training and testing set, the results are better, but surprisingly, even with photometrically modified training images, the basis functions are able to reliably handle geometric transformations. The basis functions outperform the baseline (see first row in Table~\ref{Tab:Geom}) by $\sim 50 \%$ in terms of root identification accuracy and $\sim 56\%$ in terms of IPT reconstruction accuracy. In the case of substituting the asymmetric measure in the baseline with the proposed asymmetric measure (we used Gaussian RBF), while retaining the tree spanning algorithm (see second row in Table~\ref{Tab:Geom}), an improvement of $\sim 32\%$ in terms of root identification accuracy and an improvement of $\sim 52\%$ in terms of IPT reconstruction accuracy is observed.

\begin{figure}[]
\centering

   \includegraphics[scale=.25]{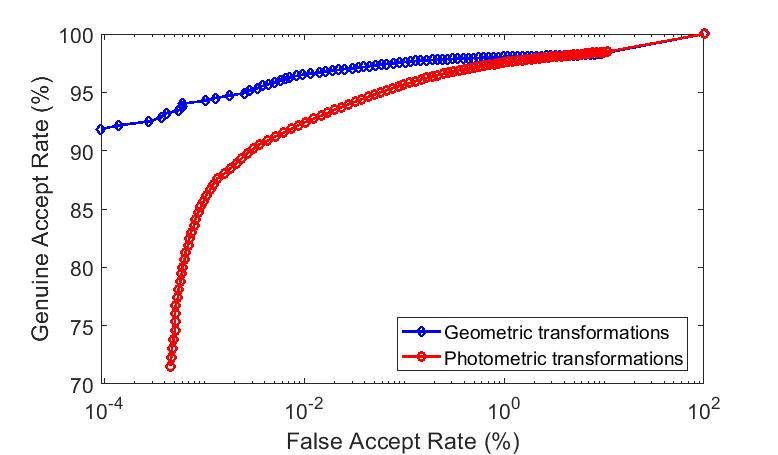}\hfill
 
\caption{Experiment 5: ROC curves for recognition of the original images with the photometrically and geometrically modified images using a COTS face matcher. The recognition performance is higher for geometrically altered images compared to photometrically modified images indicating high degree of similarity with the original images.}
\label{Fig:ROC}
\end{figure}

We also observe that the IPT reconstruction accuracy is lower for geometrically modified images compared to photometrically modified images. We tried to further analyze this difference in performance. Visual inspection revealed that the geometrically modified images appeared `more similar' to the original image compared to the photometrically altered images. This can be attributed to the restrictive parameter range used in geometric transformations. A restrictive parameter range ensures that the images are indeed near-duplicates. A wide variation in the parameter values may result in highly dissimilar images, thereby, destroying the notion of near-duplicates. To quantify the degree of similarity between the original images and the modified images, we performed face recognition using a commercial face matcher. In the face recognition experiment, the original images served as the gallery and the geometrically modified images served as the probe samples. We repeat this same process for photometrically modified images belonging to Set I of Experiment 2. ROC curves presented in Figure~\ref{Fig:ROC} indicate a true match rate of 96.27\% at a false match rate of $0.01\%$ for the geometrically altered images, and a true match rate of 91.87\% at a false match rate of $0.01\%$ for the photometrically altered images. Note, the probe and gallery sizes for the photometrically modified images are four times more than that for the geometrically modified images. Nonetheless, the basis functions can better handle the identification of the original image (root node) and reconstruction of the IPT compared to the existing method.    

\begin{table}[h]
\centering
\caption{Experiment 5: Root identification and IPT reconstruction accuracies for geometric transformations. The top two rows indicate the baseline algorithms. The baselines yield only one root node as output so results are reported only at Rank 1 and the remaining ranks are indicated Not Applicable (NA). In this experiment, the testing (TE) is always done on geometrically modified images (indicated by TE-GM) but the training (TR) can be done using either photometrically modified images (indicated by TR-PM) or geometrically modified images (TR-GM). Results indicate training on geometrically modified images yield best performance when tested on geometric transformations.}
\label{Tab:Geom}
\scalebox{0.88}{
\begin{tabular}{|llll|} \hline
\textbf{Method}               & \textbf{Protocol}                & \textbf{\begin{tabular}[c]{@{}l@{}}Root identification \\ accuracy at \\ Ranks 1/2/3 (\%)\end{tabular}} & \textbf{\begin{tabular}[c]{@{}l@{}}IPT \\ reconstruction\\ accuracy (\%)\end{tabular}} \\ \hline \hline
\multirow{2}{*}{Baseline}     & \begin{tabular}[c]{@{}l@{}}(SURF + MSAC) + \\ Oriented Kruskal \end{tabular}   & 8.00 / NA / NA                                                                                     & 3.62                                                                              \\
                              & \begin{tabular}[c]{@{}l@{}} Gaussian RBF + \\ Oriented Kruskal \end{tabular} & 27.20 / NA / NA                                                                                    & 7.24                                                                              \\ \hline
\multirow{2}{*}{Legendre}     & TR-PM, TE-GM                     & 14.20 / 24.40 / 37.00                                                                                & 52.78                                                                             \\
                              & TR-GM, TE-GM                     & 23.20 / 39.60 / 52.80                                                                                & 54.30                                                                             \\
\multirow{2}{*}{Chebyshev}    & TR-PM, TE-GM                     & 7.00 / 13.40 / 19.60                                                                                  & 51.40                                                                             \\
                              & TR-GM, TE-GM                     & 23.20 / 35.60 / 49.20                                                                                & \textbf{59.81}                                                                    \\
\multirow{2}{*}{Gabor}        & TR-PM, TE-GM                     & 25.80 / 41.20 / 52.60                                                                              & 57.75                                                                             \\
                              & TR-GM, TE-GM                     & 25.60 / 40.00 / 54.60                                                                              & 58.49                                                                             \\
\multirow{2}{*}{Gaussian RBF} & TR-PM, TE-GM                     & 25.40 / 42.40 / 57.00                                                                              & 55.67                                                                             \\
                              & TR-GM, TE-GM                     & \textbf{58.60 / 75.80 / 86.00}                                                                     & 51.40                                                                             \\
\multirow{2}{*}{Bump RBF}     & TR-PM, TE-GM                     & 7.80 / 17.60 / 31.00                                                                                   & 49.82                                                                             \\
                              & TR-GM, TE-GM                     & 46.60 / 65.80 / 77.00                                                                                  & 56.32   \\ \hline                                                                         
\end{tabular}}
\end{table}

%
\subsection{Results of Experiment 6}

\begin{figure}
\centering
\subfloat[Bob Marley near-duplicates]
{
    \includegraphics[scale=.182]{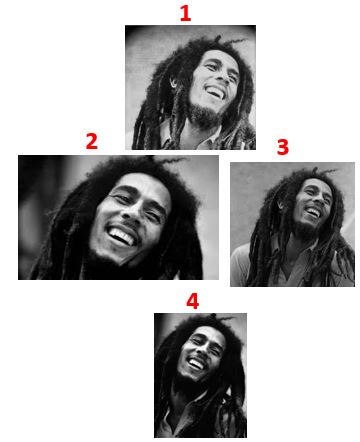}

} \hspace{0.42cm}
\subfloat[IPT constructed using Gabor]
{
    \includegraphics[scale=.182]{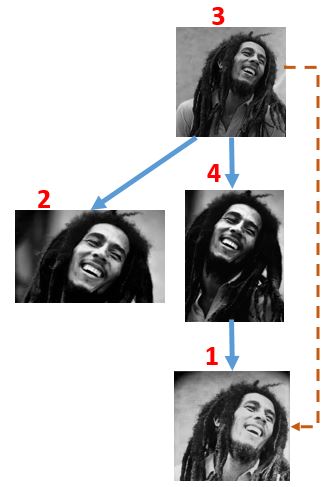}
    
} \hspace{0.42cm}
\subfloat[IPT constructed using Gaussian RBF]
{
    \includegraphics[scale=.182]{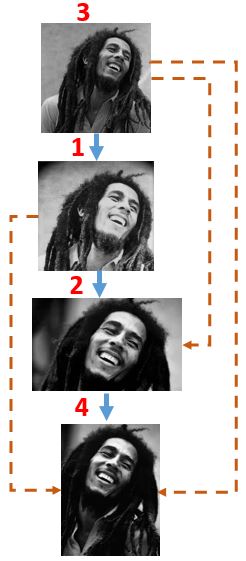}   

}\\
\subfloat[Britney Spears near-duplicates]
{
    \includegraphics[scale=.192]{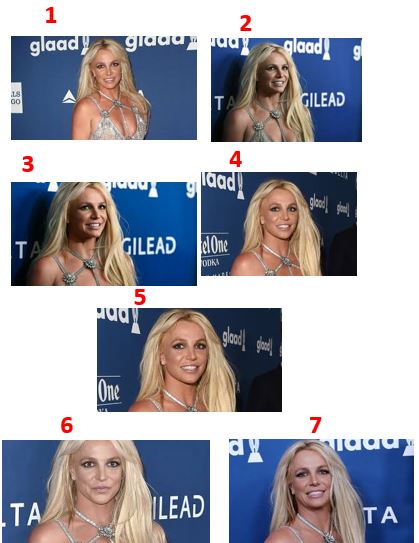}
    
}\hspace{0.7cm} 
\subfloat[IPT constructed using Chebyshev]
{
    \includegraphics[scale=.192]{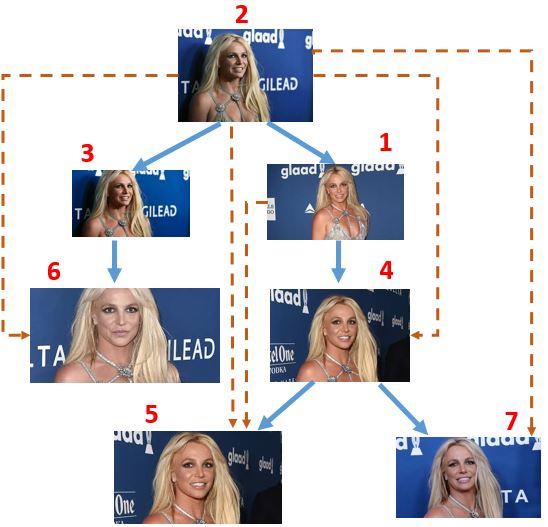}
    
}
\caption{Experiment 6: Examples of near-duplicates available online and their corresponding IPTs constructed using the proposed method. The first row corresponds to (a) 4 near-duplicates retrieved using the query \textit{Bob Marley}, (b) IPT constructed using Gabor trained on photometric distribution (the top 3 candidate root nodes are 2,3,1) and (c) IPT constructed using Gaussian RBF trained on geometric distribution (the top 3 candidate root nodes are 3,2,1). The second row corresponds to (d) 7 near-duplicates retrieved using the query \textit{Britney Spears} and (e) IPT constructed using Chebyshev trained on photometric distribution (the top 3 candidate root nodes are 2,4,5). The bold arrows indicate immediate links and the dashed arrows indicate ancestral links.}
\label{Fig: NDDR}
\end{figure}

Examples of the near-duplicates retrieved from the internet and their corresponding IPT reconstructions are presented in Figure~\ref{Fig: NDDR}. Qualitative analysis indicates that the reconstructed IPTs can depict the relationship between the near-duplicates reasonably well. For example, the images 3, 4 and 1 for the \textit{Bob Marley} images (see Figure~\ref{Fig: NDDR}(c)) should ideally follow the sequence as indicated by the IPT. Similarly, image 2 appears to be a cropped version of image 3 which is correctly constructed by the proposed method.

\subsection{Results of Experiment 7}

We used parameter distributions learnt from both photometrically modified images and geometrically modified images for IPT reconstruction. The results are reported in Table~\ref{Tab:DL} in terms of root identification and IPT reconstruction accuracies. 

\begin{figure}[h]
\centering
\subfloat[h]
{
    \includegraphics[scale=.35]{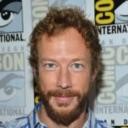} 
} \hspace{1.7cm}
\subfloat[]
{
    \includegraphics[scale=.35]{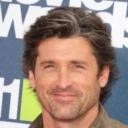} 
} \hspace{1.7cm}
\subfloat[]
{
    \includegraphics[scale=.35]{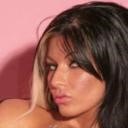} 
}
\caption{Example images from the CelebA dataset containing prominent background details in the face images. }
\label{Fig: CelAimages}
\end{figure}

1. For the near-duplicates generated using autoencoder, the Gabor model performs the best in terms of root identification accuracy (Rank 1 accuracy of 81.25\%) and the Chebyshev polynomials perform the best in terms of IPT reconstruction accuracy (64.37\%). Gabor outperforms the remaining basis functions due to the characteristics of the CelebA dataset that includes face images along with with background details (see Figure~\ref{Fig: CelAimages}). Gabor as a texture descriptor is able to accurately model the change in the texture of the background regions in the outputs reconstructed using the autoencoder.  

2. For the near-duplicates generated using the image augmentation package, maximum root identification accuracy of 50\% at Rank 3 is obtained by Bump RBF while Chebyshev obtains an IPT reconstruction accuracy of 75.20\%. The image augmentation package includes elastic distortions that are complex non-linear deformations, that the basis functions did not model accurately. 

3. For the near-duplicates generated using the BeautyGlow network, we only performed qualitative evaluation due to the small size of the test set. We observe that the IPT constructed in Figure~\ref{Fig: BG}(b) indicates a deep tree with gradual increase in the intensity of the make-up which is anticipated. 

We observed that the basis functions perform better when trained on photometrically modified images compared to when they are trained on geometrically modified images, in the case of both autoencoder and Augmentor. This is perhaps due to the fact that the autoencoder does not introduce geometric modifications; the modifications are restricted to structural and textural details. On the other hand, the image augmentation library produces random geometric modifications such as rotation and re-sampling, but uses more sophisticated techniques to remove some of the artifacts associated with such operations (\textit{e.g.}, removes the black padding near the borders of a rotated image). In such cases, we speculate that the basis functions view the geometrically altered image as a photometrically modified image. This explains the better performance of the basis functions when trained on photometrically modified images compared to geometrically modified images.

\begin{table*}[]
\centering
\caption{Experiment 7: Root identification and IPT reconstruction accuracies for deep learning-based transformations. The near duplicates are generated either using the autoencoder (left block) or image augmentation schemes for training deep neural networks (right block).}
\label{Tab:DL}
\scalebox{0.8}{
\begin{tabular}{|l|llll||llll|} \hline
                                                                                     & \multicolumn{4}{c||}{\textbf{For near-duplicates generated using autoencoder}}                                                                                                                                                                                                                                                            & \multicolumn{4}{c|}{\textbf{For near-duplicates generated using image augmentation schemes}}                                                                                                                                                                                                                                           \\
\multirow{2}{*}{\textbf{\begin{tabular}[c]{@{}l@{}}Basis \\ functions\end{tabular}}} & \multicolumn{2}{l}{\textbf{\begin{tabular}[c]{@{}l@{}}Performance when trained\\ on photometric transformations\end{tabular}}}                                      & \multicolumn{2}{l||}{\textbf{\begin{tabular}[c]{@{}l@{}}Performance when trained\\ on geometric transformations\end{tabular}}}                                       & \multicolumn{2}{l}{\textbf{\begin{tabular}[c]{@{}l@{}}Performance when trained\\ on photometric transformations\end{tabular}}}                                      & \multicolumn{2}{l|}{\textbf{\begin{tabular}[c]{@{}l@{}}Performance when trained\\ on geometric transformations\end{tabular}}}                                       \\
                                                                                     & \begin{tabular}[c]{@{}l@{}}Root identification\\ accuracy (\%) \\ @ Ranks 1/2/3\end{tabular} & \begin{tabular}[c]{@{}l@{}}IPT \\reconstruction\\ accuracy (\%)\end{tabular} & \begin{tabular}[c]{@{}l@{}}Root identification\\ accuracy (\%)\\ @ Ranks 1/2/3\end{tabular} & \begin{tabular}[c]{@{}l@{}}IPT \\reconstruction\\ accuracy (\%)\end{tabular} & \begin{tabular}[c]{@{}l@{}}Root identification\\ accuracy (\%)\\ @ Ranks 1/2/3\end{tabular} & \begin{tabular}[c]{@{}l@{}}IPT\\ reconstruction\\ accuracy (\%)\end{tabular} & \begin{tabular}[c]{@{}l@{}}Root identification\\ accuracy (\%) \\ @ Ranks 1/2/3\end{tabular} & \begin{tabular}[c]{@{}l@{}}IPT\\ reconstruction\\ accuracy (\%)\end{tabular} \\ \hline
Legendre                                                                             & 25.00 / 50.00 / 68.75                                                                       & 50.63                                                                    & 25.00 / 50.00 / 62.50                                                                       & 62.50                                                                    & 16.00 / 27.00 / 40.00                                                                       & 74.00                                                                    & 16.00 / 30.00 / 42.00                                                                       & 71.27                                                                    \\
Chebyshev                                                                            & 25.00 / 56.25 / 62.50                                                                       & \textbf{64.37}                                                           & 37.50 / 43.75 / 68.75                                                                       & 46.88                                                                    & 13.00 / 24.00 / 39.00                                                                       & \textbf{75.20}                                                           & 18.00 / 29.00 / 43.00                                                                       & 72.00                                                                    \\
Gabor                                                                                & \textbf{81.25 / 100 / 100}                                                                  & 55.00                                                                    & 31.25 / 43.75 / 56.25                                                                       & 49.38                                                                    & 12.00 / 25.00 / 36.00                                                                       & 71.93                                                                    & 18.00 / 30.00 / 40.00                                                                       & 70.27                                                                    \\
Gaussian RBF                                                                         & 87.50 / 93.75 / 93.75                                                                       & 55.00                                                                    & 43.75 / 56.25 / 68.75                                                                       & 50.62                                                                    & 17.00 / 30.00 / 43.00                                                                       & 65.20                                                                    & 19.00 / 30.11 / 41.00                                                                       & 67.67                                                                    \\
Bump RBF                                                                             & 56.25 / 68.75 / 87.50                                                                       & 57.50                                                                    & 25.00 / 37.50 / 43.75                                                                       & 55.00                                                                    & \textbf{17.00 / 32.00 / 50.00                                                                      } & 66.20                                                                    & 12.00 / 26.00 / 44.00                                                                       & 67.40  \\ \hline                                                                          
\end{tabular}}
\end{table*}

\begin{figure}
\centering

   \includegraphics[scale=.32]{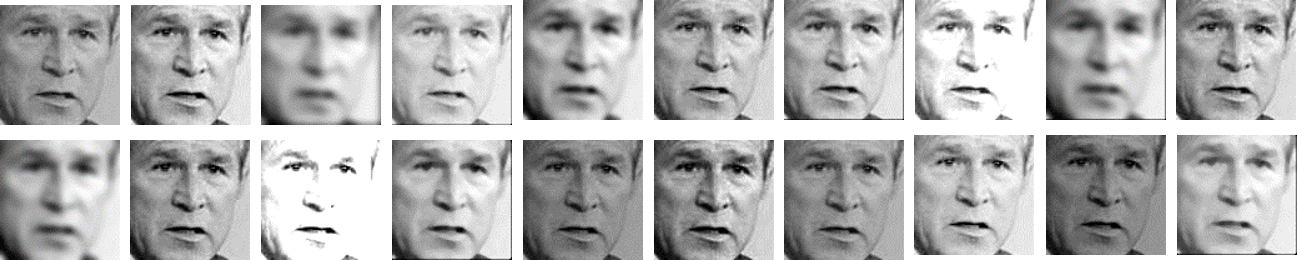}\hfill
 
\caption{An example of a set of photometrically modified images used for comparison between the proposed algorithm and human assessment. 20 such sets were simulated and each user was asked to identify their top three choices for the original image (root node). }
\label{HumanEval}
\end{figure}

In addition to the seven experiments discussed above, we performed another small experiment to compare human ability through visual inspection with the performance of the proposed algorithm. To accomplish this task, we generated 20 sets of photometrically modified images, where each set contained 20 images subjected to a random sequence of the 4 transformations described in Table~\ref{Tab1:Params}. An example of such a set is depicted in Figure~\ref{HumanEval}. Next, we asked each of 12 evaluators to indicate their top 3 preferences for the original image (root node) of each set.  

\begin{figure}
\centering

   \includegraphics[scale=.2]{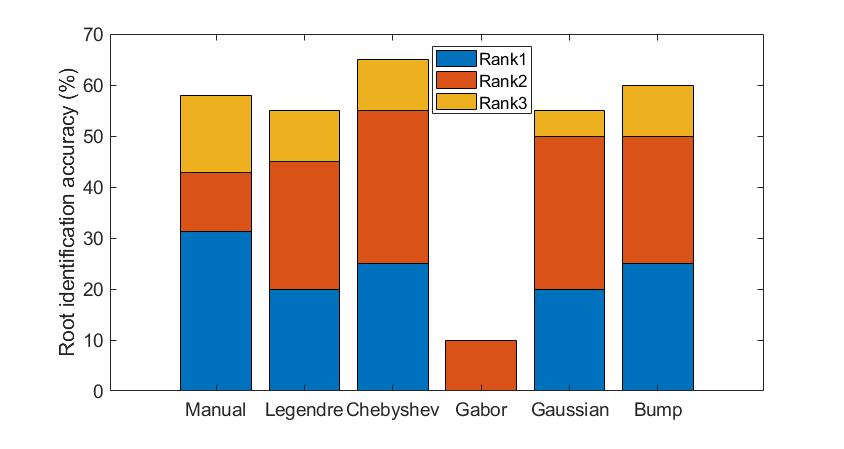}\hfill
 
\caption{Bar plots comparing the performance of human evaluators with the proposed algorithm in terms of root identification for 20 sets of images, each set containing 20 near-duplicate face images (see Figure~\ref{HumanEval}).}
\label{Bar_HumanEval}
\end{figure}

We compared the root identification accuracy as assessed by human evaluators with the output generated by the proposed algorithm for all the five basis functions employed in this work. The bar plots presented in Figure~\ref{Bar_HumanEval} indicate that the proposed algorithm performs comparably with manual evaluation. At Rank 1 the best performing algorithm (Chebyshev) is outperformed by humans by $\approx$5\%, however, at Ranks 2 and 3, the Chebyshev polynomials outperform human evaluation by $\approx$10\%.  
%
%
%

Next, we highlight the main take-away points from all the experiments.

\begin{enumerate}
\item Radial basis functions performed best at modeling the baseline photometric transformations (Brightness adjustment, Gamma transformation, Gaussian smoothing and Median filtering) and also geometric transformations. They resulted in the lowest residual error when used for modeling the transformations.
\item Orthogonal polynomials performed best at reliably discriminating between the forward and reverse directions. They resulted in the highest root identification and IPT reconstruction accuracies in a majority of the cases involving photometric and geometric modifications. 
\item The proposed approach of utilizing ``likelihood ratio" generalizes well across multiple IPT configurations and different biometric modalities.
\item The proposed approach is capable of handling different classes of transformations both photometric and geometric. In addition, they generalize well over previously unseen transformations based on deep learning tools and image editing software.
\end{enumerate}

\begin{figure}[h]
\centering
\subfloat[]
{
    \includegraphics[scale=.23]{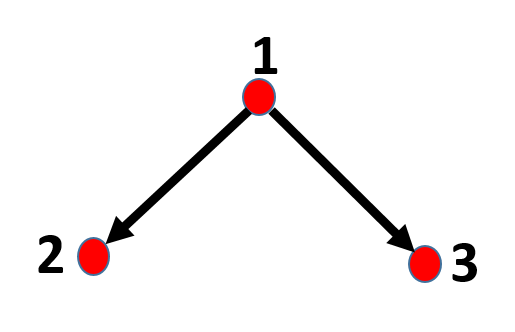} 
} \hspace{0.5cm}
\subfloat[]
{
    \includegraphics[scale=.23]{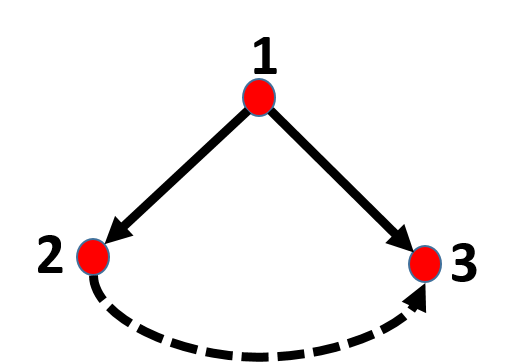} 
} \hspace{0.5cm}
\subfloat[]
{
    \includegraphics[scale=.23]{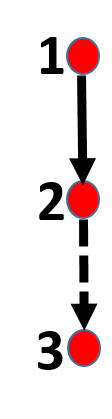} 
}

\caption{Toy example demonstrating the effect of inserting a spurious edge on the von Neumann entropy. (a) Groundtruth IPT (b) Correctly reconstructed IPT with spurious edge (c) Incorrectly reconstructed IPT with spurious edge. Note, the spurious edge is indicated by a dashed line.}
\label{Ent_Demo}
\end{figure}

\begin{table}[]
\centering
\caption{Approximate von Neumann entropy for analysis of spurious edges. The mean and the standard deviation of the differences between the entropy of the ground truth and the reconstructions are reported. Low values indicate accurate reconstructions and smaller number of spurious as well as missing edges.}
\label{Tab:Entropy}
\scalebox{0.9}{
\begin{tabular}{|l|l|} \hline
\textbf{Basis Function} & \textbf{\begin{tabular}[c]{@{}l@{}}Entropy (Mean\\ and standard deviation)\end{tabular}} \\ \hline
Legendre6               & -0.0009 $\pm$ 0.0045                                                                         \\
Chebyshev               & \textbf{-0.0000 $\pm$ 0.0044}                                                                \\
Gabor                   & -0.0036 $\pm$ 0.0039                                                                         \\
Gaussian RBF            & -0.0035 $\pm$ 0.0028                                                                         \\
Bump                    & -0.0034 $\pm$ 0.0030                                                                        \\ \hline
\end{tabular}}
\end{table}

\section{Explanatory Model}
\label{Sec:Explan}
Finally, in this section, we present a graph entropy-based explanatory model which analyzes the failure cases of the IPTs reconstructed using the proposed method. A failed IPT reconstruction can involve (i) missing edges, and (ii) spurious edges. Such failure cases can be quantified using the approximate von Neumann entropy measure for directed graphs. Graph entropy is computed in terms of its in-degree and out-degree~\cite{Entropy}. Consider a directed graph $G(V,E)$ with a set of nodes denoted by $V$ and the set of edges denoted by $E$. The adjacency matrix of such a graph is defined as,

 \tab \tab \tab $A_{uv} = \begin{cases} 1 &\mbox{if } (u,v) \in E, \\ 
0 & \mbox{otherwise.}  \end{cases}$

\noindent The in-degree and out-degree of node $u$ are presented as $\displaystyle d^{in}_u=\sum_{v\in V}A_{vu};   d^{out}_u=\sum_{v\in V}A_{uv}$. The von Neumann entropy for the directed graph $G$ can be written as: 
\begin{equation} 
\label{Eqn:entropy}
H = 1-\frac{1}{|V|}-\frac{1}{2|V|^2}\bigg[\sum_{(u,v)\in E}\frac{d^{in}_u}{d^{in}_v d^{out^2}_u}\!+\!\sum_{(u,v)\in E_2}\frac{1}{d^{out}_u d^{out}_v }\bigg]. 
\end{equation}
Here, $E_1=\{(u,v) |(u,v) \in E$ and $(v,u)\notin E\}$; $E_2=\{(u,v)|(u,v)\in E$ and $(v,u)\in E\}$, such that, $E=E_1\cup E_2$ and $E_1 \cap E_2=\emptyset $. 
The maximum value of entropy for a directed graph is equal to $1-\frac{1}{|V|}$ for a star graph where all the nodes in the graph have either incoming links or outgoing links but not both. The minimum entropy is obtained for a cyclic graph, implying all nodes are fully connected, and the value of minimum entropy is $1- \frac{1}{|V|}- \frac{1}{2|V|^2}|V| = 1- \frac{1}{|V|}- \frac{1}{2|V|}=1-\frac{1.5}{|V|}$. In our work, we want to move toward maximum entropy since minimum entropy results in the worst case scenario of reconstruction where the result is a cyclic graph. Thus, Eqn.~(\ref{Eqn:entropy}) can be simplified as follows. 
\begin{equation} 
\label{Eqn:entropy_reduced}
H = 1- \frac{1}{|V|} - \frac{1}{2|V|^2} \bigg[\sum_{(u, v) \in E_1} \frac{d^{in}_u}{d^{in}_v d^{out^2}_u} \bigg]. 
\end{equation}
The above equation is applicable in our work since $E_2 = \emptyset$, otherwise, we will end up having a cycle in the reconstructed IPT. We hypothesize that spurious edges decrease the entropy of the IPT, which is further reduced if the IPT misses correct edges. This can be illustrated using a toy example presented in Figure~\ref{Ent_Demo}. The reconstruction yields a 100\% IPT reconstruction accuracy for the first reconstructed IPT (Figure~\ref{Ent_Demo}(b)), which has no missing edge but one spurious edge, and 50\% for the second reconstructed IPT (Figure~\ref{Ent_Demo}(c)), which has one missing edge and one spurious edge. The von Neumann entropy as computed from Eqn.~(\ref{Eqn:entropy_reduced}) for the ground truth configuration (Figure~\ref{Ent_Demo}(a)) is,
$\displaystyle H(GT)\!=\!1\!-\!\frac{1}{3}\!-\!\frac{1}{18}\!\bigg[\!\frac{d^{in}_1}{d^{in}_2 d^{out^2}_1}+\!\frac{d^{in}_1}{d^{in}_3 d^{out^2}_1}\bigg]\!=\!1\!-\!\frac{1}{3}\!-\!\frac{1}{18}\bigg[0\bigg]\!=\!0.67$. Note that the in-degree of node 1 is 0. This value also corresponds to the maximum entropy of a directed graph with 3 nodes. The entropy for the first reconstructed IPT (Figure~\ref{Ent_Demo}(b)) is, $\displaystyle H(IPT1)\!=\!1\!-\!\frac{1}{3}\!-\!\frac{1}{18}\!\bigg[\frac{d^{in}_1}{d^{in}_2 d^{out^2}_1}+\frac{d^{in}_1}{d^{in}_3 d^{out^2}_1}+\frac{d^{in}_2}{d^{in}_3 d^{out^2}_2}\bigg]\!=\!1\!-\!\frac{1}{3}\!-\!\frac{1}{18}\bigg[0\!+\!0\!+\!\frac{1}{2\times\!(1)^2}\bigg]=\!0.64$. The entropy for the second reconstructed IPT (Figure~\ref{Ent_Demo}(c)) is, $\displaystyle H(IPT2)\!=\!1\!-\!\frac{1}{3}\!-\!\frac{1}{18}\bigg[\frac{d^{in}_1}{d^{in}_2 d^{out^2}_1}\!+\!\frac{d^{in}_2}{d^{in}_3 d^{out^2}_2}\bigg]\!=\!1\!-\!\frac{1}{3}\!-\!\frac{1}{18}\bigg[0\!+\!\frac{1}{1\times\!(1)^2}\!\bigg]\!=\!0.61$. Thus, $H(GT)>H(IPT1)>H(IPT2)$. This demonstrates that inaccurate IPTs (missing edges and spurious edges) can reduce the entropy from the ground truth entropy, and this property can be leveraged for evaluating the goodness of the reconstructed IPTs. Reconstructed IPTs with missing edges and spurious edges will tend to have lower entropy.

We compute this entropy-based measure to analyze the accuracy of the IPTs reconstructed using the proposed method for the face images (full set with 2,727 IPTs). The maximum and minimum entropy for a graph containing 10 nodes is 0.85 and 0.90, respectively. The von Neumann entropy as computed from Eqn.~(\ref{Eqn:entropy_reduced}) for the ground truth IPT (see Figure~\ref{Fig:TestIPTs_ALL}(a)) is 0.89. 
We then compute the entropy corresponding to the reconstructed IPTs. Finally, we compute the difference between the entropy of the ground truth configurations and the reconstructed configurations. We report the mean and the standard deviation of the differences in the entropy in Table~\ref{Tab:Entropy}. The results indicate that the differences between the ground truth entropy and the entropy of the reconstructed IPTs are the smallest when the polynomials are utilized as basis functions. This further corroborates our findings that the polynomials result in the best IPT reconstruction accuracies (see Tables~\ref{Tab:Face_Set1} - \ref{Tab:IntraModality}). 

\section{Conclusion}
\label{Sec:Concl}
In this work, we presented a method to model pairwise photometric transformations between a set of near-duplicate biometric images using basis functions. The modeling of the photometric transformations is used in conjunction with the likelihood ratio based asymmetric measure to identify the original image and deduce how the images are related to each other. We performed comprehensive experiments using face, iris and fingerprint images under different experimental settings including intra-modality, cross-modality, and cross-dataset scenarios. Results indicate promising performance with a root identification accuracy of $\approx 94\%$ and an IPT reconstruction accuracy of $\approx 72\%$. We also analyzed the performance of the basis functions using $t$-SNE and observed that some of the photometric transformations are easier to model than others. We also successfully modeled geometric transformations using the basis functions. We observed that the proposed scheme can be generalized to previously unseen transformations accomplished using image editing software and deep learning-based methods. Finally, we utilized graph entropy measure to evaluate the reconstructed IPTs. The deviations between the ground truth and reconstructed IPTs are minimal indicating that the proposed reconstruction algorithm performs reasonably well.         
Future work will focus on performing a global analysis that will simultaneously consider all the nodes to derive accurate IPT reconstructions. 

\section{Acknowledgments}
This material is based upon the work supported by the National Science Foundation under Grant Number 1618518.

\balance
\small
\bibliographystyle{ieeetran}
\bibliography{TBIOM}

\begin{thebibliography}{10}
\providecommand{\url}[1]{#1}
\csname url@samestyle\endcsname
\providecommand{\newblock}{\relax}
\providecommand{\bibinfo}[2]{#2}
\providecommand{\BIBentrySTDinterwordspacing}{\spaceskip=0pt\relax}
\providecommand{\BIBentryALTinterwordstretchfactor}{4}
\providecommand{\BIBentryALTinterwordspacing}{\spaceskip=\fontdimen2\font plus
\BIBentryALTinterwordstretchfactor\fontdimen3\font minus
  \fontdimen4\font\relax}
\providecommand{\BIBforeignlanguage}[2]{{%
\expandafter\ifx\csname l@#1\endcsname\relax
\typeout{** WARNING: IEEEtran.bst: No hyphenation pattern has been}%
\typeout{** loaded for the language `#1'. Using the pattern for}%
\typeout{** the default language instead.}%
\else
\language=\csname l@#1\endcsname
\fi
#2}}
\providecommand{\BIBdecl}{\relax}
\BIBdecl

\bibitem{Face}
V.~{\v S}truc and N.~Pavesic, ``Photometric normalization techniques for
  illumination invariance,'' in \emph{Advances in Face Image Analysis:
  Techniques and Technologies}.\hskip 1em plus 0.5em minus 0.4em\relax
  IGI-Global, 01 2011, pp. 279--300.

\bibitem{Face_tampering}
A.~Agarwal, A.~Sehwag, R.~Singh, and M.~Vatsa, ``Deceiving face presentation
  attack detection via image transforms,'' \emph{IEEE International Conference
  on Multimedia Big Data}, 08 2019.

\bibitem{Farid_08_DIF}
H.~Farid, ``Digital image forensics,'' \emph{Scientific American}, vol. 298,
  no.~6, pp. 66--71, 2008.

\bibitem{Ross_ICB_19}
A.~Ross, S.~Banerjee, C.~Chen, A.~Chowdhury, V.~Mirjalili, R.~Sharma,
  T.~Swearingen, and S.~Yadav, ``Some research problems in biometrics: The
  future beckons,'' in \emph{12th {IAPR} International Conference on Biometrics
  ({ICB})}, Crete, Greece, June 2019.

\bibitem{ImageForensics_FBI}
N.~A. {Spaun}, ``Forensic biometrics from images and video at the federal
  bureau of investigation,'' in \emph{IEEE International Conference on
  Biometrics: Theory, Applications, and Systems ({BTAS})}, Sep. 2007, pp. 1--3.

\bibitem{Chainofcustody_Ref2}
J.~{\'C}osi{\'c} and M.~Ba{\v c}a, ``({I}m)proving chain of custody and digital
  evidence integrity with time stamp,'' in \emph{The 33rd International
  Convention MIPRO}, May 2010, pp. 1226--1230.

\bibitem{ChainofCustody_FP}
N.~{Bartlow}, N.~{Kalka}, B.~{Cukic}, and A.~{Ross}, ``Identifying sensors from
  fingerprint images,'' in \emph{IEEE Computer Society Conference on Computer
  Vision and Pattern Recognition Workshops}, June 2009, pp. 78--84.

\bibitem{Moreira_18_Provenance}
D.~Moreira, A.~Bharati, J.~Brogan, A.~da~Silva~Pinto, M.~Parowski, K.~W.
  Bowyer, P.~J. Flynn, A.~Rocha, and W.~J. Scheirer, ``Image provenance
  analysis at scale,'' \emph{IEEE Transactions on Image Processing (T-IP)},
  vol.~27, no.~12, 2018.

\bibitem{Bharati_17_UPhylogeny}
A.~Bharati, D.~Moreira, A.~Pinto, J.~Brogan, K.~Bowyer, P.~Flynn, W.~Scheirer,
  and A.~Rocha, ``U-phylogeny: Undirected provenance graph construction in the
  wild,'' in \emph{IEEE International Conference on Image Processing (ICIP)},
  05 2017.

\bibitem{Dias_12_MST}
Z.~Dias, A.~Rocha, and S.~Goldenstein, ``Image phylogeny by minimal spanning
  trees,'' \emph{IEEE Transactions on Information Forensics and Security},
  vol.~7, no.~2, pp. 774--788, April 2012.

\bibitem{Dias_13_largescaleIPT}
Z.~Dias, S.~Goldenstein, and A.~Rocha, ``Large-scale image phylogeny: Tracing
  image ancestral relationships,'' \emph{IEEE MultiMedia}, vol.~20, no.~3, pp.
  58--70, July 2013.

\bibitem{Dias_13_OB}
------, ``Exploring heuristic and optimum branching algorithms for image
  phylogeny,'' \emph{Journal of Visual Communication and Image Representation},
  vol.~24, no.~7, pp. 1124 -- 1134, 2013.

\bibitem{Bestagini_16_regionbasedIPT}
P.~Bestagini, M.~Tagliasacchi, and S.~Tubaro, ``Image phylogeny tree
  reconstruction based on region selection,'' in \emph{IEEE International
  Conference on Acoustics, Speech and Signal Processing (ICASSP)}, March 2016,
  pp. 2059--2063.

\bibitem{Philippe_16_missingmarkersIPT}
N.~L. Philippe, W.~Puech, and C.~Fiorio, ``Phylogeny of {JPEG} images by
  ancestor estimation using missing markers on image pairs,'' in \emph{Sixth
  International Conference on Image Processing Theory, Tools and Applications
  (IPTA)}, Dec 2016, pp. 1--6.

\bibitem{Milani_16_agemetricsIPT}
S.~Milani, M.~Fontana, P.~Bestagini, and S.~Tubaro, ``Phylogenetic analysis of
  near-duplicate images using processing age metrics,'' in \emph{IEEE
  International Conference on Acoustics, Speech and Signal Processing
  (ICASSP)}, March 2016, pp. 2054--2058.

\bibitem{Dias_13_AutoIPF}
Z.~Dias, S.~Goldenstein, and A.~Rocha, ``Toward image phylogeny forests:
  Automatically recovering semantically similar image relationships,''
  \emph{Forensic Science International}, vol. 231, no. 1–3, pp. 178 -- 189,
  2013.

\bibitem{Oliveira_14_multipleparentingIPT}
A.~Oliveira, P.~Ferrara, A.~D. Rosa, A.~Piva, M.~Barni, S.~Goldenstein,
  Z.~Dias, and A.~Rocha, ``Multiple parenting identification in image
  phylogeny,'' in \emph{IEEE International Conference on Image Processing
  (ICIP)}, Oct 2014, pp. 5347--5351.

\bibitem{SpecClus_16_Dias}
M.~A. Oikawa, Z.~Dias, A.~de~Rezende~Rocha, and S.~Goldenstein, ``Manifold
  learning and spectral clustering for image phylogeny forests,'' \emph{IEEE
  Transactions on Information Forensics and Security}, vol.~11, no.~1, pp.
  5--18, Jan 2016.

\bibitem{Melloni_14_dissmetricsIPT}
A.~Melloni, P.~Bestagini, S.~Milani, M.~Tagliasacchi, A.~Rocha, and S.~Tubaro,
  ``Image phylogeny through dissimilarity metrics fusion,'' in \emph{Fifth
  European Workshop on Visual Information Processing (EUVIP)}, Dec 2014, pp.
  1--6.

\bibitem{Costa_17}
F.~Costa, A.~Oliveira, P.~Ferrara, Z.~Dias, S.~Goldenstein, and A.~Rocha, ``New
  dissimilarity measures for image phylogeny reconstruction,'' \emph{Pattern
  Analysis and Applications}, vol.~20, 03 2017.

\bibitem{Ban_17_IJCB}
S.~Banerjee and A.~Ross, ``Computing an image phylogeny tree from
  photometrically modified iris images,'' in \emph{Proc. of 3rd International
  Joint Conference on Biometrics (IJCB)}, October 2017.

\bibitem{FacePhyloTree_2019}
------, ``Face phylogeny tree: Deducing relationships between near-duplicate
  face images using legendre polynomials and radial basis functions,'' in
  \emph{10th IEEE International Conference on Biometrics: Theory, Applications
  and Systems ({BTAS}), Tampa, Florida}, September 2019.

\bibitem{Leg1}
S.~Omachi and M.~Omachi, ``Fast template matching with polynomials,''
  \emph{IEEE Transactions on Image Processing}, vol.~16, no.~8, pp. 2139--2149,
  Aug 2007.

\bibitem{Leg2}
G.~Li and C.~Wen, ``Legendre polynomials in signal reconstruction and
  compression,'' in \emph{5th IEEE Conference on Industrial Electronics and
  Applications}, June 2010, pp. 1636--1640.

\bibitem{GCN_NIPS_16}
M.~Defferrard, X.~Bresson, and P.~Vandergheynst, ``Convolutional neural
  networks on graphs with fast localized spectral filtering,'' in
  \emph{Proceedings of the 30th International Conference on Neural Information
  Processing Systems}, 2016, pp. 3844--3852.

\bibitem{Bartoli_08_Photomodel}
A.~Bartoli, ``Groupwise geometric and photometric direct image registration,''
  \emph{IEEE Transactions on Pattern Analysis and Machine Intelligence},
  vol.~30, no.~12, pp. 2098--2108, Dec 2008.

\bibitem{Baker_04_IJCV}
S.~Baker and I.~Matthews, ``Lucas-{K}anade 20 years on: A unifying framework,''
  \emph{International Journal of Computer Vision}, vol.~56, no.~3, pp.
  221--255, 2004.

\bibitem{Daug_Gabor}
J.~G. Daugman, ``Uncertainty relation for resolution in space, spatial
  frequency, and orientation optimized by two-dimensional visual cortical
  filters,'' in \emph{Journal of Optical Society of America A}, vol.~2, no.~7,
  July 1985, pp. 1160--1169.

\bibitem{Bump}
L.~W. Tu, ``Bump functions and partitions of unity,'' in \emph{An Introduction
  to Manifolds}.\hskip 1em plus 0.5em minus 0.4em\relax New York, NY: Springer
  New York, 2008, pp. 127--134.

\bibitem{KDE}
C.~Sammut, ``Density estimation,'' in \emph{Encyclopedia of Machine Learning
  and Data Mining}, C.~Sammut and G.~I. Webb, Eds.\hskip 1em plus 0.5em minus
  0.4em\relax Boston, MA: Springer US, 2017, pp. 348--349.

\bibitem{LFWTech}
G.~B. Huang, M.~Ramesh, T.~Berg, and E.~Learned-Miller, ``Labeled faces in the
  wild: A database for studying face recognition in unconstrained
  environments,'' University of Massachusetts, Amherst, Tech. Rep. 07-49,
  October 2007.

\bibitem{CasV2}
``C{ASIA} {I}ris {D}atabase {V}ersion 2,'' \url{
  http://biometrics.idealtest.org/dbDetailForUser.do?id=2}, [Online accessed:
  12th April 2019].

\bibitem{CASv4}
``C{ASIA} {I}ris {D}atabase {V}ersion 4,'' \url{
  http://biometrics.idealtest.org/dbDetailForUser.do?id=4}, [Online accessed:
  30th August 2019].

\bibitem{FVC}
D.~{Maio}, D.~{Maltoni}, R.~{Cappelli}, J.~L. {Wayman}, and A.~K. {Jain},
  ``{FVC}2000: fingerprint verification competition,'' \emph{IEEE Transactions
  on Pattern Analysis and Machine Intelligence}, vol.~24, no.~3, pp. 402--412,
  March 2002.

\bibitem{TSNE}
L.~van~der Maaten and G.~E. Hinton, ``Visualizing data using t-{SNE},''
  \emph{Journal of Machine Learning Research}, vol.~9, pp. 2431--2556, November
  2008.

\bibitem{NDDR}
J.~J. Foo, J.~Zobel, and R.~Sinha, ``Clustering near-duplicate images in large
  collections,'' in \emph{Proceedings of the International Workshop on Workshop
  on Multimedia Information Retrieval}.\hskip 1em plus 0.5em minus 0.4em\relax
  New York, NY, USA: Association for Computing Machinery, 2007, p. 21–30.

\bibitem{CAE}
X.~Guo, X.~Liu, E.~Zhu, and J.~Yin, ``Deep clustering with convolutional
  autoencoders,'' in \emph{Neural Information Processing}, D.~Liu, S.~Xie,
  Y.~Li, D.~Zhao, and E.-S.~M. El-Alfy, Eds.\hskip 1em plus 0.5em minus
  0.4em\relax Cham: Springer International Publishing, 2017, pp. 373--382.

\bibitem{Augmentor}
``Augmentor: Image augmentation library in python for machine learning,''
  \url{https://github.com/mdbloice/Augmentor}, [Online accessed: 3rd Januray,
  2020].

\bibitem{CelebA}
``Celeb{A} dataset,'' \url{http://mmlab.ie.cuhk.edu.hk/projects/CelebA.html},
  [Online accessed: 3rd January, 2020].

\bibitem{BeautyGlow}
H.-J. Chen, K.-M. Hui, S.-Y. Wang, L.-W. Tsao, H.-H. Shuai, and W.-H. Cheng,
  ``Beauty{G}low: On-demand makeup transfer framework with reversible
  generative network,'' in \emph{The IEEE Conference on Computer Vision and
  Pattern Recognition (CVPR)}, June 2019.

\bibitem{Entropy}
C.~Ye, R.~C. Wilson, C.~H. Comin, L.~da~F.~Costa, and E.~R. Hancock, ``Entropy
  and heterogeneity measures for directed graphs,'' in \emph{Similarity-Based
  Pattern Recognition}, E.~Hancock and M.~Pelillo, Eds.\hskip 1em plus 0.5em
  minus 0.4em\relax Berlin, Heidelberg: Springer Berlin Heidelberg, 2013, pp.
  219--234.

\end{thebibliography}

\begin{IEEEbiography}[{\includegraphics[width=0.9in,height=0.95in,clip, keepaspectratio]{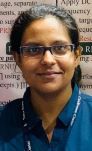}}]{Sudipta Banerjee}
received the B.Tech. degree in Electronics and Communication Engineering from  West Bengal University of Technology, India, in 2011, and the M.E. degree in Electronics and Tele-Communication Engineering with specialization in Control Engineering from Jadavpur University, India, in 2014. She is currently pursuing her Ph.D. degree in Computer Science and Engineering at Michigan State University under the supervision of Dr. Arun Ross in the iPRoBe research laboratory. Her research interests include image forensics in the context of digitally altered biometric images.
\end{IEEEbiography}

\begin{IEEEbiography}[{\includegraphics[width=0.9in,height=0.95in,clip,keepaspectratio]{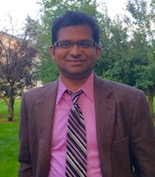}}]{Arun Ross}
is the John and Eva Cillag Endowed Chair in the College of Engineering and a Professor in the Department of  Computer Science and Engineering at Michigan State University, United States. He received the B.E. (Hons.) degree in Computer Science from BITS Pilani, India, and the M.S. and PhD degrees in Computer Science and Engineering from Michigan State University, United States. He served as the faculty at West Virginia University between 2003 and 2012 where he received the Benedum Distinguished Scholar  Award for excellence in creative research and the WVU Foundation Outstanding Teaching Award. Ross was a recipient of  the NSF CAREER Award and was designated a Kavli Fellow by the US National Academy of Sciences in 2006. He received the JK Aggarwal Prize in 2014 and the Young Biometrics Investigator Award in 2013 from the International Association of  Pattern Recognition. He is the co-author of the textbook, “Introduction to Biometrics”and the monograph, “Handbook of Multibiometrics”.
\end{IEEEbiography}

\end{document}